\documentclass[sigconf,screen]{acmart} 

\AtBeginDocument{%
  }

\renewcommand\footnotetextcopyrightpermission[1]{}

\usepackage{multirow}
\usepackage{enumitem}
\usepackage{makecell}
\usepackage{booktabs}
\usepackage{framed}
\usepackage{bbding}
\newtheorem{definition}{Definition}
\usepackage{amsmath,amsfonts}
\usepackage{subfig}
\usepackage[title]{appendix}
\usepackage{algorithm}
\usepackage{algorithmic}

\begin{document}

\title{Primary and Secondary Factor Consistency as Domain Knowledge to Guide Happiness Computing in Online Assessment}


\author{Xiaohua Wu}
\email{xhwu@whut.edu.cn}
\affiliation{%
  \institution{Wuhan University of Technology}
  \country{China}
  \postcode{430070}
}

\author{Lin Li}
\email{cathylilin@whut.edu.cn}
\affiliation{%
	\institution{Wuhan University of Technology}
	\country{China}
	\postcode{430070}
}

\author{Xiaohui Tao}
\email{xiaohui.tao@unisq.edu.au}
\affiliation{%
	\institution{University of Southern Queensland}
	\country{Australia}
}

\author{Frank Xing}
\email{xing@nus.edu.sg}
\affiliation{%
	\institution{National University of Singapore}
	\country{Singapore}
}

\author{Jingling Yuan}
\email{yjl@whut.edu.cn}
\affiliation{%
	\institution{Wuhan University of Technology}
	\country{China}
	\postcode{430070}
}

\renewcommand{\shortauthors}{Xiaohua Wu, et al.}

\begin{abstract}
	Happiness computing based on large-scale online web data and machine learning methods is an emerging research topic that underpins a range of issues, from personal growth to social stability. Many advanced Machine Learning (ML) models with explanations are used to compute the happiness online assessment while maintaining high accuracy of results. However, domain knowledge constraints, such as the primary and secondary relations of happiness factors, are absent from these models, which limits the association between computing results and the right reasons for why they occurred. This article attempts to provide new insights into the explanation consistency from an empirical study perspective. Then we study how to represent and introduce domain knowledge constraints to make ML models more trustworthy. We achieve this through: (1) proving that multiple prediction models with additive factor attributions will have the desirable property of primary and secondary relations consistency, and (2) showing that factor relations with quantity can be represented as an importance distribution for encoding domain knowledge. Factor explanation difference is penalized by the Kullback-Leibler divergence-based loss among computing models. Experimental results using two online web datasets show that domain knowledge of stable factor relations exists. Using this knowledge not only improves happiness computing accuracy but also reveals more significative happiness factors for assisting decisions well.
\end{abstract}


\maketitle

\section{Introduction}
With the rapid development of mobile and web technologies, online social networks and survey data furnish rich information on human's happiness concerns and help government leaders, politicians, and public figures understand the needs and aspirations of their citizens. This information improves personal growth, health, and the overall development of society by enabling appropriate decisions for strategic urban planning \cite{Musa_2018,Bublyk2022}. Ed Diener, a famous psychologist whose research focused on theories and measurements of well-being, argued that economic criteria alone cannot evaluate the state of our society, further stating that national indicators of happiness must also be included as a basis for policy decisions \cite{Randy_2008}. Unraveling what constitutes happiness is therefore of both theoretical and practical significance. \citet{Layard_2005} indicated that happiness levels can be increasingly used as a measure of economic development, and even as a primary measure.

Identifying the key happiness factors for citizen groups based on online web data has become a topic of great interest in recent years. To assist in decision-making, some studies \cite{Randy_2008,Wanting_2013} conclude that happiness is stable for specific groups, since it is a combination of both long-term emotional and life satisfaction, although there are various breakdowns of happiness factors. Stemming from this, \citet{Kahneman_2006} defines two main types of factors affecting happiness level: (1) a current emotion such as pleasure or joy, and (2) the quality of life over a period of time. Therefore, there are shared primary happiness factors, especially within a demographic group.

\begin{figure*}[ht]
	\centering
	\includegraphics[width=0.92\textwidth]{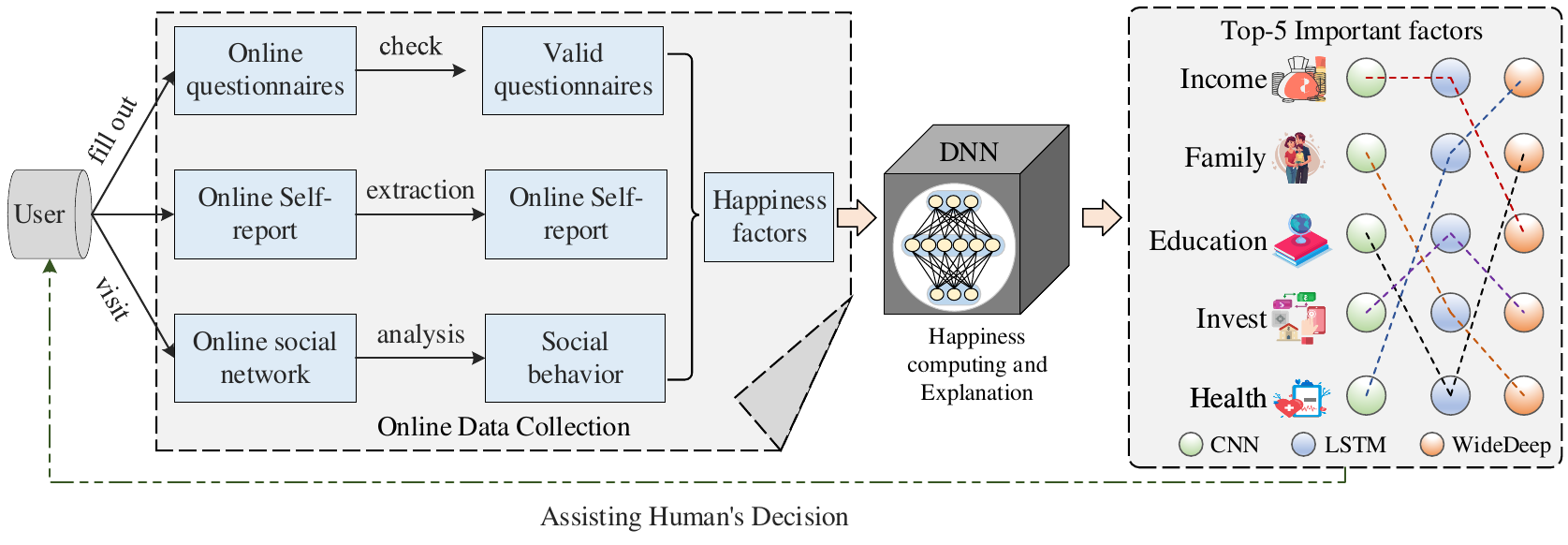}
	\caption{The illustration of our research motivation. The various online web data is collected from users online, then is applied to the happiness computing models and the factor explanation is generated by an explanation method. }
	\label{Task_Define}
\end{figure*}

\begin{figure}[t]
	\centering
	\subfloat[primary factors]{
		\includegraphics[width=0.49\linewidth]{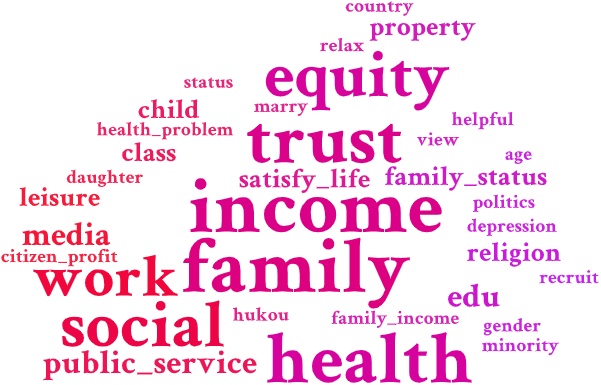}
		\label{top10}
	}
	\subfloat[secondary factors]{
		\includegraphics[width=0.49\linewidth]{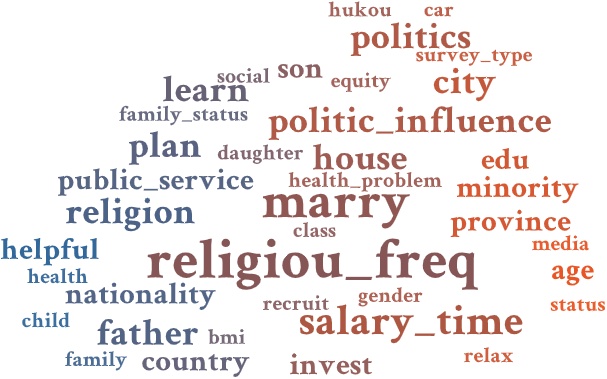}
		\label{last10}
	}
	\caption{The primary factors and secondary factors for happiness level. }
	\label{domain_knowledge_sample}
\end{figure}

Knowing the primary happiness factors is even more important than improving the prediction accuracies to support decision-making in our society. This leads to a preference for using simple yet highly interpretable modeling methods, such as probabilistic models, regression analysis, and other conventional Machine Learning (ML) methods, rather than complex models over a period of time. For example, numerous regression analysis methods are employed to research the relationship between happiness and factors such as mental health \cite{Garaigordobil_2015,Wang_Weiwei_2022}, income \cite{Oshio_2010,Makridis_2021,Wang_Weiwei_2022}, emotion and depression \cite{Kahriz_2020}, social support, and physical activity \cite{Dadvand_2016}. However, these models are less accurate for the growing prevalence of large-scale social media and online questionnaires. To mitigate this issue, recent studies have introduced deep neural networks (DNN) in happiness computing. A follow-up effort by \cite{Shao_2021} leverages DNN to understand the impact of three key demographic variables (age, gender, and race) using the impact and interaction scores. Additionally, convolutional neural network (CNN), bidirectional long short-term memory with Attention (BiLSTMA), WideDeep, and pre-trained models (e.g., BERT, happyBERT, and ELMo) have been utilized to compute the happiness level \cite{Rajendran_2019,Evensen_2019,Sweeney_2022}. Nevertheless, these deep learning models are usually a ``black box'' and supervised only by happiness level signs, which usually makes happiness computing models predict right not for the right reasons and thereby damages the decision-making well.

Taking the happiness computing and factor explanation as an example, in Figure \ref{Task_Define}, from the online collected data, the happiness computing models (CNN, LSTM and WideDeep) are explained by a post-hoc explanation method. The factor relations of different models are various, so which one can be applied in practice?

Therefore, there is a pressing need to constrain a model's training by leveraging more signs.
To detect, and thus be able to correct such behavior, advances have been made in using a model's causal explanation as guidance for enhancing model training \cite{Lopez_Paz_2017,Cui_Peng_2021,Selvaraju_2020,Shrikumar_2017}. The aim of this approach is to constrain the models to be right for the right reasons \cite{Ross_2017}. However, these studies focus only on individual happiness level or factor explanation to train the models, which is affected by the case-by-cause results.

Motivated by the latest works that penalize explanations using prior knowledge \cite{RiegerSMY2020,Schramowski_2020}, we compute the factor importance on different groups and sort the importance, then the intersection of top-$k$ factors and last-$k$ factors on these groups are produced respectively displayed in Figure~\ref{domain_knowledge_sample}. We find that there are primary and secondary relations among the happiness factors. Therefore, we try to utilize these relations as domain knowledge to improve happiness computing. In response, the two fundamental, non-trivial challenges need to be solved:

\begin{itemize}[leftmargin=*]
	\item \textbf{\textit{How to represent the happiness domain knowledge?}} As constructed in social science research, there are many types of domain knowledge such as factor distribution, and primary and secondary factor relations. However, it is difficult to determine these representations.

	\item \textbf{\textit{How to guide model training using domain knowledge?}} While happiness computing models have shown promising results, a lack of domain knowledge may lead to inconsistent computing results among different models guided only by labels. It is thus unable to guide the models to compute the right happiness level for the right reasons.
\end{itemize}

To this end, we posit that primary and secondary factor relation consistency is an effective domain knowledge representation. The computing models can be constrained by using domain knowledge and happiness level to optimize for correct outputs, as well as for attributions to human primary factors relations. The main contributions of this paper are summarized as follows:
\begin{enumerate}[leftmargin=*]
	\item We provide empirical evidence demonstrating that there is factor explanation difference among popular happiness computing models by the same explanation methods and same datasets;
	\item We theoretically prove that the multiple computing models with additive factor attribution have the desirable property of primary and secondary factor relation consistency.
	The relations with quantity are represented as importance distributions for encoding domain knowledge, which is used to constrain the models' training;
	\item This novel method is implemented on four specific groups in the online web data Chinese General Social Survey (CGSS) and the European Social Survey (ESS). Our results verify the improvements in terms of accuracy, with reliability further confirmed through a theoretical and practical implication of  online happiness assessment.
\end{enumerate}

\section{Related Work}
\label{Related Work}

\subsection{Happiness Factor Analysis}
How to find the right reasons for happiness prediction is a key research question, one that first received significant attention in sociological research. Studies from this domain found that happiness is primarily associated with income, health, family, and more factors, via regression analysis and ML ~\cite{Richard_2001,Saputri_2015,Yu_2017,Laaksonen_2018}. These conventional ML-based prediction methods are interpretable, but are generally unable to reliably model the complex relationships and achieve a stable accuracy performance. This gap jeopardizes the usability of ML models for quality, high-level decision-making.

Increasingly, with the development of DNN, some researchers proposed various DNN-based methods to analyze the relationship between factors and happiness, although these works offer limited interpretability \cite{Perez-BenitoVCG19,Xin_Inkpen_2019,Islam_Goldwasser_Yoga_Happy_2020}. To address these problems, a follow-up effort by \cite{Shao_2021} used DNN to understand the impact of three key demographic variables (age, gender, and race) using the impact and interaction scores. A study by \cite{ijcai2022_Lilin} also aimed to find key factors through quantitative analysis, providing explanations via Attention and Shapley value. From the comparison of explanation results, they found that there are indeed primary and secondary factors for happiness level. Nonetheless, these models are supervised by happiness level and lack of explanation for constraints, which usually cannot achieve the right prediction for the right reasons.

\subsection{Model Constraint Using Explanations}
Training guidance by a happiness level signal may usually lead a model to obtain the spurious correlations between factors and happiness level in the training data, which may result in the right prediction, but not necessarily for the right reasons. Therefore, besides only labels, model training with explanation guidance has come to the attention of current researchers \cite{Godbole_2004}. Increasingly, studies have begun to use the model explanation to assist in supervising the training process \cite{Yuanqing_2021,cikm_WangLLZ2022,wacv_WatsonHM23}. Human explanations are employed to ensure models focus on relevant features and prevent them from fitting to spurious correlations in the data \cite{Teso_2019,Cui_Peng_2021}. Additional follow-up approaches for constraining DNN have also been proposed, such as domain knowledge \cite{Bolukbasi_2016}, projecting out superficial statistics \cite{Wang_2019}, etc. Although these models are improved by the guidance of their explanation, the explanation results are case-by-case and are based on individual dataset, task, or model, which is not robust for application to other areas. Moreover, these explanations are usually based on Attention or model weight, both of which are less theoretically supported \cite{Sarthak_2019,Serrano_2019}. It is noted that these studies are in computer science, a field that is parallel to this work. In a word, the gold standard for any ML model is to be able to achieve high levels of accuracy whilst learning the concrete reasons produced in the data.

\section{Empirical Analysis}
\subsection{Online Happiness Assessment}
The happiness computing in online assessment has cast as a classification problem depend on the online web datasets, where given an instance $x^{(n)}=\{x^{(n)}_1, x^{(n)}_2, \dots, x^{(n)}_j\}$, $x^{(n)} \in \mathcal{X}$ and $x^{(n)}_j$ represents the factor $j$ of instance $x^{(n)}$, the objective is to compute the happiness level $y_{ic}$ from the candidate levels $\mathcal{Y}$. This computation is based on the collected instances from online web. Without loss of generality, a happiness computing model is can be formulated as a function transformation $F: \mathcal{X} \mapsto \mathcal{Y}$. The objective function $p(.)$ is formulated as:
\begin{equation}
	\hat{y_{ic}} = \underset{y_{ic} \in \mathcal{Y}}{arg max} \;  p(y|x; \Theta),
	\label{equ:task_define}
\end{equation}
where $\Theta$ denotes the model parameters. The classification loss $\mathcal{L}_{label}$ is defined as Equation~\ref{cross_entropy}.

\subsection{Revisiting Question in Happiness Computing}

\subsubsection{Are There Explanation Difference?}

In online happiness assessment, various models are usually employed on the collected online data. From the recent work, we find that the explanation results by multiple models are usually different in view of happiness factor importance. To attempt to verify that there are explanation differences in happiness computing, to begin, we conduct the popular happiness computing models on two large-scale online web datasets. Then, a post-hoc explanation method is employed for computing a unified factor importance and investigating the explanation consistency. These experiment settings are as follows.

\paragraph{Dataset} We conduct comprehensive experiments to evaluate the performance of our method on the specific groups of the public Chinese General Social Survey (CGSS)\footnote{http://cgss.ruc.edu.cn/} 2015 (full edition) and the European Social Survey (ESS)\footnote{https://ess-search.nsd.no/} 2018 datasets.

Specifically, the CGSS is an open shared and large-scale social online investigation dataset, with the subject being Chinese families containing more than 8000 samples and 124 factors per sample. The ESS is an academically-driven, multi-country online web survey that includes 50k samples and 102 factors per one over 38 countries to date. It has been extensively used to effectively assess the progress of nations and develop a series of European social indicators around citizens' happiness and well-being. Many studies in sociology, economics, and even politics have utilized ESS data. It is noted that happiness is scaled by 5 levels (1-5), where a higher level indicates more happiness. There are more than 124 factors in CGSS and 102 factors in ESS. The overview of these two datasets is presented on Table~\ref{dataset_descirbe}. More factor details of these two datasets are presented in \textit{Appendix \ref{Appendix:Factors_Introduction}}.

The groups are usually split by some basic factors, which can cut the groups with discrepancy as much as possible. For example, we may categorize the age groups by standard young ($\le$40) and elder ($>$40) measures, which is a common practice in China \cite{Peng_2004,Xue_2011}. The health status is defined by the National Survey Research Center of China and European Research Infrastructure Consortium in the original questionnaires for the datasets, and is split into health bad (1-3) and health good (4-5). Our solution can be extended to other happiness online web data such as happyDB \cite{Akari_2018}.

\begin{table}[t] \small
	\centering
	\tabcolsep=0.5mm
	\renewcommand\arraystretch{1.05}
	\caption{The statistics of CGSS and ESS datasets.}
	\label{dataset_descirbe}
	\begin{tabular}{cccccc}
		\toprule
		\multicolumn{3}{c}{CGSS} & \multicolumn{3}{c}{ESS} \\
		\cmidrule(r){1-3}		\cmidrule(r){4-6}
		\makecell{Happiness Level} & sample & factor & Happiness Level & sample & factor \\
		\midrule
		1 & 77 & 124 & 1 & 721 & 102 \\
		2 & 315 & 124 & 2 & 1515 & 102 \\
		3 & 630 & 124 & 3 & 3991 & 102 \\
		4 & 1743 & 124 & 4 & 6453 & 102 \\
		5 & 422 & 124 & 5 & 2877 & 102 	\\ \bottomrule
	\end{tabular}
\end{table}

\begin{figure}[t]
	\centering
	\includegraphics[width=1.0\linewidth]{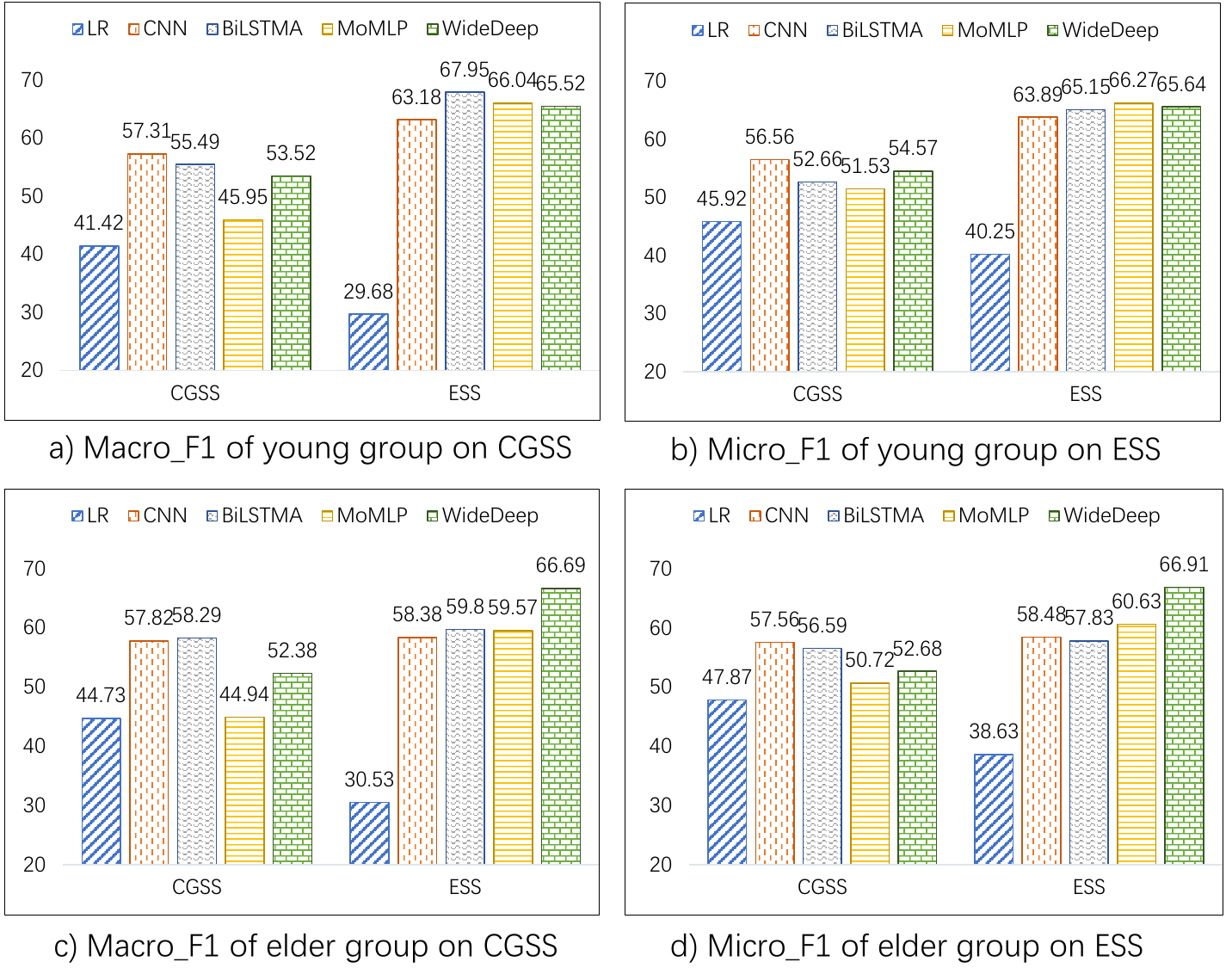} 
	\caption{The Macro\_F1 and Micro\_F1 results of young group and elder group on CGSS and ESS datasets. }
	\label{F1_before_DK}
\end{figure}

\paragraph{Base Happiness Computing Models} As discussed in the introduction, there are two categories of happiness computing models considered in our experiments. The first is the conventional ML methods consisting of Logistics Regression (LR) \cite{Yu_2017}, and multi-output MLP (MoMLP) \cite{Park_2020}. The second is DNN-based methods, including CNN \cite{Omid_Student_Happiness_2019}, BiLSTMA \cite{Islam_Goldwasser_Yoga_Happy_2020}, and WideDeep \cite{ijcai2022_Lilin}. The WideDeep is a jointly trained wide linear and deep model that can combine the benefits of memorization and generalization, and possesses a strong ability to extract multiple-factor interactions. 
The details of our baselines are listed as follows:
\begin{itemize}[leftmargin=*]
	\item \textit{Logistics Regression (LR)}: An outstanding algorithm that has both abilities of classification and interpretability.
	\item \textit{Multi-output MLP (MoMLP)}: The MLP is configured in such a way that the output layer comprises multiple output neurons, and each output neuron indicates the probability of belonging to the class it represents.
	\item \textit{Convolutional Neural Network (CNN)}: The neural network exhibits great performance in prediction tasks. We build a three-layer network structure with one-dimensional convolution.
	\item \textit{BLSTM with Attention (BiLSTMA)}: A popular model that can measure the long-term dependence of factors. The BiLSTMA is based on the LSTM network and attention mechanism. The LSTM is equipped with 128 hidden units and 2 layers.
	\item \textit{WideDeep (WD)}: A jointly trained wide linear and deep model that can combine the benefits of memorization and generalization. It's employed due to the strong ability to extract multiple-factor interaction.
\end{itemize}

\paragraph{Parameter Settings and Metrics} In addition, to compute the factor importance based on happiness prediction models, the SHAP \cite{Lundberg_2017} is introduced in this work for efficient Shapley value \cite{Shapley_2016} calculation. We implement the experiments in Python 3.7.0 with PyTorch 1.7.1 on a commodity server equipped with 256G memory and an Intel E5-2650 CPU. We train these models by using Adam optimizer with momentum of 0.9, weight decay of 1e-4, and set batch\_size to 128. The learning rate is 1e-4, the manual seed is 101 for the critical aspect of reproduction, and loss function $\mathcal{L}$ is defined as Equation \ref{final_loss}. The source code will be available after publication.

In this work, Macro\_F1 and Micro\_F1 are used for a multi-classification model through \textit{k}-fold cross-validation ($k$=5) \cite{ijcai2022_Lilin}. For the evaluation of accuracy stability, the shorter the file box, the better the stability. To evaluate the factor relation consistency, Kendall's tau coefficient is employed. It is defined as Equation \ref{kendall_tau}.
\begin{equation}
	\tau = \frac{\mathcal{C} - \mathcal{D}}{n(n-1)/2},
	\label{kendall_tau}
\end{equation}
where $\mathcal{C}$ represents the number of concordant pairs, and $\mathcal{D}$ is the number of discordant pairs. $n$ is the number of ranked factors in each column. The higher the value of Kendall's tau coefficient, the better the factor relation consistency.

\paragraph{Verified Results} The experimental results on young and elder group of CGSS and ESS datasets are presented in Figure \ref{F1_before_DK}. It obviously shows that there are competitive Macro\_F1 and Micro\_F1 accuracies in two groups, especially among models such as CNN, BiLSTMA, MoMLP, and WideDeep. However, does the comparable accuracy mean the consistent explanation results?

\begin{figure}[t]
	\centering
	\subfloat[\textit{young} group of CGSS]{
		\includegraphics[width=0.49\linewidth]{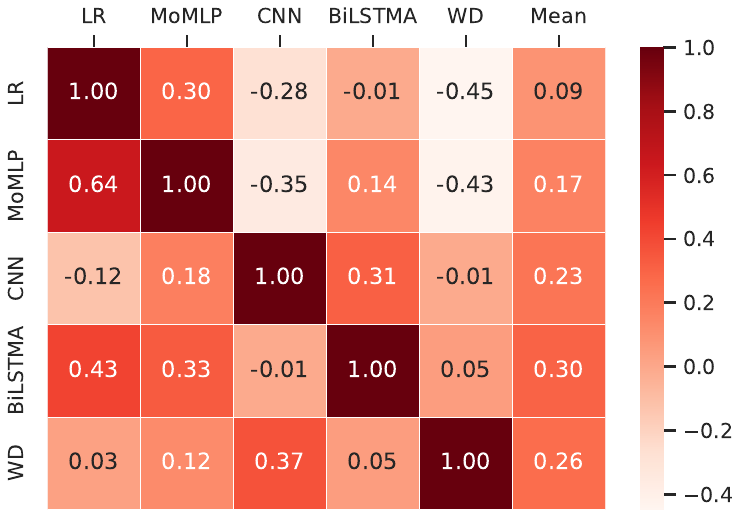}
	}
	\subfloat[\textit{young} group of ESS]{
		\includegraphics[width=0.49\linewidth]{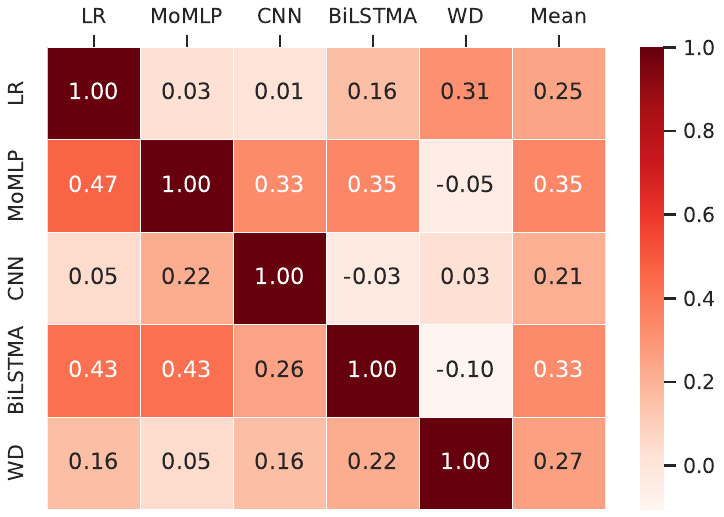}
	} \\
	\subfloat[\textit{elder} group of CGSS]{
		\includegraphics[width=0.49\linewidth]{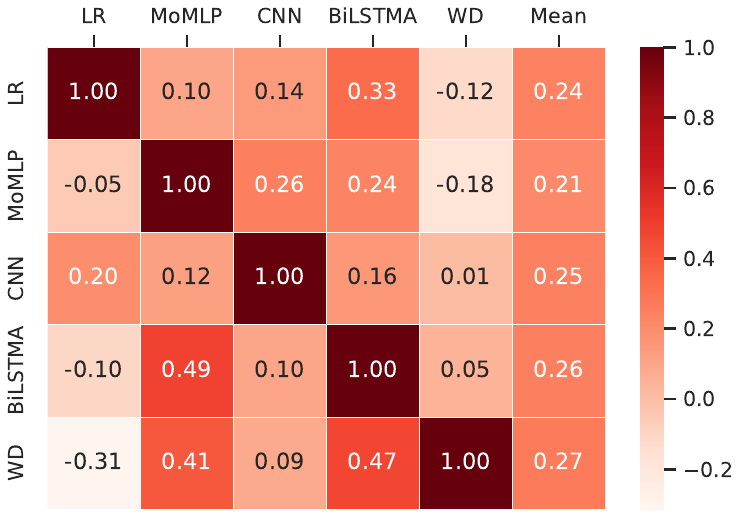}
	}
	\subfloat[\textit{elder} group of ESS]{
		\includegraphics[width=0.49\linewidth]{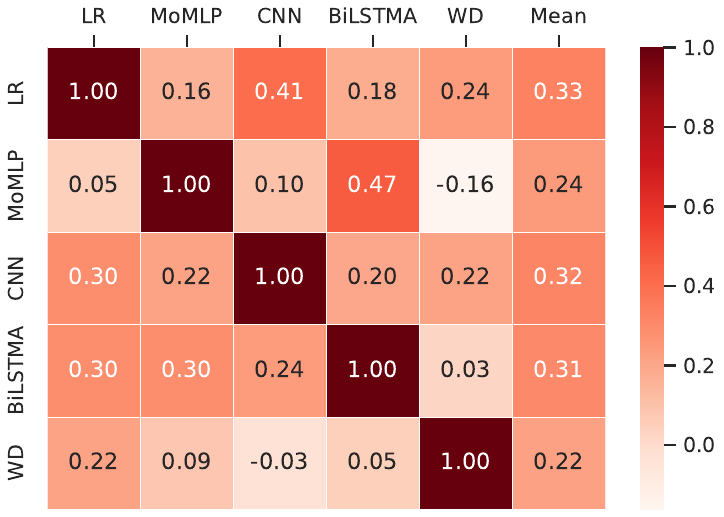}
	}
	\caption{The comparison of explanation consistency (Kendall's tau coefficient) based on the factor distribution on \textit{young} and \textit{elder} groups of CGSS and ESS datasets.}
	\label{kendall_tau_result_before_DK}
\end{figure}

\subsubsection{Explanation Difference}

In response, the factor explanation based on these models is shown in Figure~\ref{kendall_tau_result_before_DK}. The explanation consistency is lower than 0.3 evaluated by Kendall's tau coefficient on \texttt{young} and \texttt{elder} groups, which indicates that there is low consistency among multiple happiness computing models. Specifically, on the \texttt{young} group of CGSS dataset, the best results appear at BiLSTMA and only up to 0.3, but the lowest one is 0.09. Moreover, for the elder group in the CGSS and ESS datasets,  we can draw the same conclusion as the young group.

In summary, these results reveal that the explanation is different even if they have been computed by one explanation method and evaluated by one metric, which could be due to the complex and spurious correlations between factors and happiness \cite{Cui_Peng_2021}.  Since the low consistency, we are difficult to accept which one supports our making decision.

\section{How to Improve Explanation Consistency?}

Inspired by previous works, we utilize domain knowledge (e.g., factor importance, primary and secondary factor relations) to constrain the online happiness assessment and improve the explanation consistency for assisting decision-making. However, this raised two non-trivial questions: \textit{how to represent the knowledge} and \textit{how to guide model training}. As countermeasures, in this section, we first present the domain knowledge representation, i.e., the importance computing and the primary and secondary factor relations in specific groups. Second, we discuss how the knowledge is employed to guide the models' training for the right prediction for the right reasons.

\subsection{Domain Knowledge: Additive Factor Attribution}
To gain ensemble knowledge with broadly acceptable factor importance for guiding training, various ML models are considered to vote for the final results. However, the factor importance produced by the discrepant mechanisms of these methods cannot be summarized directly. To combat this, the additive property should be a fundamental condition for the explanation methods in our task. Specifically, the property is defined below.

\begin{definition} \label{definition1}
	\textit{\textbf{Additive factor attribution}. The importance of each factor can be summarized through the factor's contribution to more than one model, e.g.,
		\begin{equation}
			g(\mathcal{X}) = \phi_{0} + {\textstyle \sum_{j=1}^{J} \phi_{j}},
		\end{equation}
		where $\mathcal{X}$ denotes the survey sample with all present factors, and $g$ is an explanation model. $J$ is the number of factors, and $\phi_{j}$ is the importance of factor $j$.}
\end{definition}

In this sense, the computing models can be explained by Shapley value \cite{Shapley_2016}, which is an additive explanation method that can satisfy this definition with a theoretical guarantee. A fair explanation for a happiness factor contribution can be provided owing to its remarkable properties like efficiency, symmetry, dummy, and linearity. In comparison to other additive factor attribution methods, such as LIME \cite{Ribeiro2016} and DeepLIFT \cite{Shrikumar_2017}, Shapley value has a unique approach that satisfies the accuracy, missingness, and consistency properties of feature attribution. Therefore, we measure the pairwise proximity via similarity, which includes all factor importance produced by Shapley value. The broadly acceptable factor importance distribution can balance the explanation for all models.

\subsubsection{Importance Computing}

We consider happiness computing as a cooperative task with the ultimate goal of accurately computing the happiness level of a group with all present factors. Therefore, the importance of a factor is defined as $\phi_{j}$.
The $\phi_{j}$ is defined as follows \cite{Shapley_2016}:
\begin{equation}
	\begin{split}
		\phi_{j}(v) =&  {\textstyle \sum\limits_{S\subseteq J \setminus \left \{ x_j \right \}}\frac{|S|!(|J|-|S|-1)!}{|J|!} \left[v(S\cup  \left \{ x_j \right \}) - v(S) \right]},     \label{Shapley}
	\end{split}
\end{equation}
where $j=1,...,|J|$, $J = \left \{ x_1,\cdots, x_{|J|} \right \}$ and $S\subseteq J \setminus \left \{ x_j \right \} $ denotes all possible subsets of factor set $J$, which excludes the factor $j$ and consists of $|S|$ factors. $\frac{|S|!(J-|S|-1)!}{J!}$ is the possibility of a subset $S$. $v(S\cup \left \{ x_j \right \})- v(S)$ means the marginal contribution of factor $j$ where $v(x)\in \mathbb{R}$ denotes the model output when variables in $S$ is present. In other words, the gain is a weighted average over contribution function difference in all subsets $S$, excluding the factor $j$.

For the four desirable properties of Shapley values, i.e., efficiency, symmetry, dummy, and linearity, it is beneficial to get trustworthy explanation results \cite{Shapley_2016,Young_1985}. The details of these properties are defined in \textit{Appendix \ref{Appendix:Properties_of_Shapley}.}

\subsubsection{Consistent Attribution} Based on the properties of efficiency and symmetry, our prediction model $f(x)$ can be explained by Shapley value as indicated above. Specifically, in a specific factor set $\mathcal{M}$, the happiness level computed through models is fully voted by the contribution of all factors, which can produce the final factor importance representation.
In addition, the same contribution of two different happiness factors gives the same explanation based on the property of \textit{symmetry} and is for consistent and trustworthy explanations. Therefore, this is a natural requirement for the explanation methods in our task.

There is a principle of monotonicity based on the properties of \textit{dummy} and \textit{linearity}, which denotes that if the prediction changes so that factor contributions to all factor coalitions increase or stay the same, then the factor's allocation should not decrease \cite{Young_1985}. It is beneficial to the representation of importance by additive factor attribution, and is supplemental evidence for the primary and secondary factor relation consistency.

Owing to these four properties, the quantitative explanations for happiness factors can be fairly produced, which is beneficial in evaluating their importance. However, it is difficult to exactly calculate them, as they are exponential in the size of factors. Consequently, SHAP (SHapley Additive exPlanations) \cite{Lundberg_2017} can be utilized for efficient computation of factor importance.

\begin{table*} 
	\centering
	\tabcolsep=0.7mm
	\renewcommand\arraystretch{1.1}
	\caption{The results of all methods with or without happiness domain knowledge on \textit{young}, \textit{elder}, \textit{health good} and \textit{health bad} groups of CGSS and ESS datasets. The results of \textit{health good} and \textit{health bad} groups are presented in \textit{Appendix \ref{Appendix:F1_Results_health}}}
	\label{macro_micro_f1}
	\begin{tabular}{ccccccccccccc}
		\toprule
		\multirow{2}{*}{Group}       & \multirow{2}{*}{Dataset} & \multicolumn{1}{c}{\multirow{2}{*}{\makecell{Domain \\ Knowledge}}} & \multicolumn{2}{c}{CNN[2019]}       & \multicolumn{2}{c}{LR[2015]}      & \multicolumn{2}{c}{BiLSTMA[2020]}                                      & \multicolumn{2}{c}{MoMLP[2020]}                                       & \multicolumn{2}{c}{WideDeep [2022]}  \\
		\cmidrule(r){4-5}		 \cmidrule(r){6-7}		 \cmidrule(r){8-9}	 \cmidrule(r){10-11}  \cmidrule(r){12-13}
		&                          & \multicolumn{1}{c}{}                                                  & \multicolumn{1}{c}{Macro\_F1} & \multicolumn{1}{c}{Micro\_F1} & \multicolumn{1}{c}{Macro\_F1} & \multicolumn{1}{c}{Micro\_F1} & \multicolumn{1}{c}{Macro\_F1} & \multicolumn{1}{c}{Micro\_F1} & \multicolumn{1}{c}{Macro\_F1} & \multicolumn{1}{c}{Micro\_F1} & \multicolumn{1}{c}{Macro\_F1} & \multicolumn{1}{c}{Micro\_F1}  \\
		\midrule
		\multirow{4}{*}{Young}       & \multirow{2}{*}{CGSS}    & \XSolidBrush
		& 57.31 & 56.56 & 41.42 & 45.92 & 55.49 & 52.66 & 45.95 & 51.53 & 53.52 & 54.57 \\
		&   	& \checkmark
		& \textbf{58.26} & \textbf{57.68} & \textbf{46.27} & \textbf{49.78} & \textbf{56.95} & \textbf{54.37} & \textbf{47.38} & \textbf{53.34} & \textbf{57.16} & \textbf{56.94} \\
		& \multirow{2}{*}{ESS}     & \XSolidBrush
		& 63.18 & 63.89 & 29.68 & 40.25 & 67.95 & 65.15 & 66.04 & 66.27 & 65.52 & 65.64 \\
		&   	& \checkmark
		& \textbf{65.26} & \textbf{64.17} & \textbf{30.95} & \textbf{41.18} & \textbf{68.50} & \textbf{65.75} & \textbf{67.42} & \textbf{67.57} & \textbf{67.03} & \textbf{67.10} \\
		\multirow{4}{*}{Elder}       & \multirow{2}{*}{CGSS}    & \XSolidBrush
		& 57.82 & \textbf{57.56} & 44.73 & 47.87 & 58.29 & 56.59 & 44.94 & 50.72 & 52.38 & 52.68 \\
		&   	& \checkmark
		& \textbf{58.20} & 56.63 & \textbf{46.04} & \textbf{49.19} & \textbf{59.07} & \textbf{57.28} & \textbf{46.40} & \textbf{51.34} & \textbf{59.71} & \textbf{59.43} \\
		& \multirow{2}{*}{ESS}     & \XSolidBrush
		& 58.38 & \textbf{58.48} & 30.53 & \textbf{38.63} & 59.80 & 57.83 & 59.57 & 60.63 & 66.69 & 66.91 \\
		&   	& \checkmark
		& \textbf{58.51} & 58.40 & \textbf{30.64} & 38.59 & \textbf{60.30} & \textbf{58.45} & \textbf{60.35} & \textbf{61.32} & \textbf{67.18} & \textbf{67.18} \\
		\bottomrule
	\end{tabular}
\end{table*}

\subsection{Domain Knowledge: Primary and Secondary Factor Relation Consistency}
In the above section, we have presented the factor importance by additive factor attribution, but this is still not enough to guide the model training and to get the right prediction results for the right reasons, as we discussed in our related work. Inspired by social science study \cite{Lundberg_2017} and our experimental results shown in Figure~\ref{domain_knowledge_sample}, we find that there is primary and secondary factor relation consistency in one happiness computing model. In this section, we prove that there is the same domain knowledge of factor relation consistency among multiple models by additive factor attribution as defined in Definition \ref{definition1}. Therefore, the details of primary and secondary factor relation consistency are as follows.

\textbf{Property 1} \textit{Given an explanation function $\phi: \mathcal{X}^m \mapsto \mathbb{R}$, and a factor set $\mathcal{M}=\{x^{(n)}_1, x^{(n)}_2, \cdots, x^{(n)}_m\}$ of sample $x^{(n)}$, $S \subseteq \mathcal{M}$, $S \setminus \{i\}$ denotes setting factor $i = 0$. If for any two happiness computing models $f_k$ and $f_l$, }

\begin{equation}
	f_k(\mathcal{S}) - f_k(\mathcal{S} \setminus \{i\}) \ge f_l(\mathcal{S}) - f_l(\mathcal{S} \setminus \{i\})
\end{equation}
for all input $\mathcal{S}$, then $\phi_i(f_k, x^{(n)}_i) \ge \phi_i(f_l, x^{(n)}_i)$.

From Property 1, we infer that the factor relations is relatively consistent among different happiness computing models. Therefore, the combination of multiple models is more stable and consistent than the individual.

\textbf{Property 2}  \textit{Now we assume that $\mathcal{S}_{pri}$ and $\mathcal{S}_{sec}$ are primary factors and secondary factors, respectively, and assume that $\phi_i(f_k, x^{(n)}_i) \ge \phi_j(f_k, x^{(n)}_j)$, $i \in \mathcal{S}_{pri}$ and $j \in \mathcal{S}_{sec}$, then $\phi_i(f_l, x^{(n)}_i) \ge \phi_j(f_l, x^{(n)}_j)$. Additionally, since property dummy and linearity hold on, therefore, }

\begin{equation}
	\textstyle \sum_{k=1}^{K} \phi_i(f_k, x^{(n)}_i) \ge  \textstyle \sum_{k=1}^{K} \phi_j(f_k, x^{(n)}_j),
\end{equation}
where $\phi_i(f_{k}, x^{(n)}_i)$ denotes the Shapley value of factor $i$ based on model $f_k$.

Let us consider both Properties 1 and 2. We can infer that the primary and secondary factors relations are stable based on the importance computed by Shapley value. Therefore, the happiness domain knowledge (importance and factor relations) can be embedded by the Shapley value, which is then beneficial to guide the model training.

\subsection{How to Guide Model Training?}

In the above, we have represented the happiness domain knowledge, but it is also crucial that leveraging this knowledge guide the model's training. To mitigate this problem, the factor relations with quantity are represented as global importance distribution for encoding domain knowledge. Second, model training is implemented by the domain knowledge to achieve better happiness prediction accuracy.

\subsubsection{Global Importance Distribution}
There are $|F|$ factor importance distributions, which are produced by model $f \in F$. Therefore,  the combined factor importance is produced by Equation~\ref{consistent_shap_value}.
\begin{equation}
	Exp(x) = \frac{1}{|F|} \sum\nolimits_{f=1}^{|F|} \lambda_f exp_f,
	\label{consistent_shap_value}
\end{equation}
where $exp_f$ represents the Shapley value of all happiness factors based on model $f$, and $\lambda_f$ denotes a balancing hyper-parameter, which is the prediction accuracy of model $f$. $|F|$ indicates the number of selected models.

\subsubsection{Model Training Guidance}
Following prior work, our happiness prediction models are trained for a $c$ level classification task. The classification loss $\mathcal{L}_{label}$ is defined as Equation~\ref{cross_entropy}.
\begin{equation}
	\mathcal{L}_{label} = - \frac{1}{N} \sum_{i=1}^{N} \sum_{c=1}^{C} y_{ic} log(\hat{y}_{ic}),
	\label{cross_entropy}
\end{equation}
where the $N$ and $C$ are the number of samples and happiness levels, respectively. $y_{ic}$ is the true label in $ \{1,\cdots,C\}$ and $\hat{y}_{ic}$ denotes the predictive probability of level $c$ in sample $i$. Additionally, the Euclidean distance without considering variances cannot represent the probabilistic similarity between probability distributions and KL divergence can evaluate the relative entropy information gain \cite{Park_S22}. Therefore, the Kullback-Leibler Divergence is employed as explanation loss $\mathcal{L}_{exp}$, which is defined as Equation~\ref{KL}.

\begin{equation}
	\mathcal{L}_{exp} = \sum_{k,p=1}^{|F|} log(softmax(\phi_{f_k})) log \frac{log(softmax(\phi_{f_k}))}{softmax(\phi_{f_p})},
	\label{KL}
\end{equation}
where $\phi_{f_k}$ and $\phi_{f_p}$ denote the importance of happiness computing models $f_k, f_p \in F$, and $|F|$ is the total number of selected happiness models. Therefore, the joint training loss is defined as Equation~\ref{final_loss}.
\begin{equation}
	\min \limits_{\Theta}	\mathcal{L} = \mathcal{L}_{label} + \lambda_f \mathcal{L}_{exp},
	\label{final_loss}
\end{equation}
where the $\lambda_f$ denotes the accuracy of model $f$. It is employed as a "residual ratio" to help control the trade-off between two terms for better performance.

\begin{figure*}[ht]
	\centering
	\subfloat[\textit{young on CGSS}]{
		\includegraphics[width=0.24\linewidth]{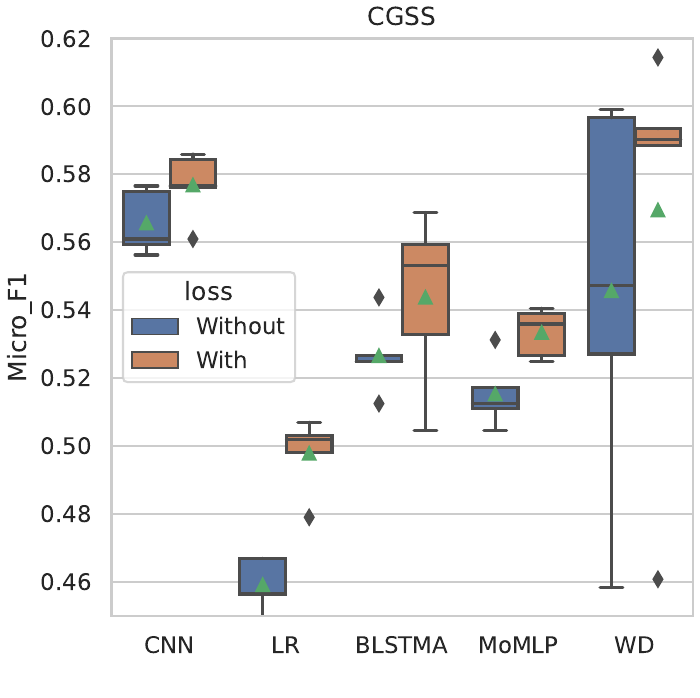} \label{young}
	}
	\subfloat[\textit{elder on CGSS}]{
		\includegraphics[width=0.24\linewidth]{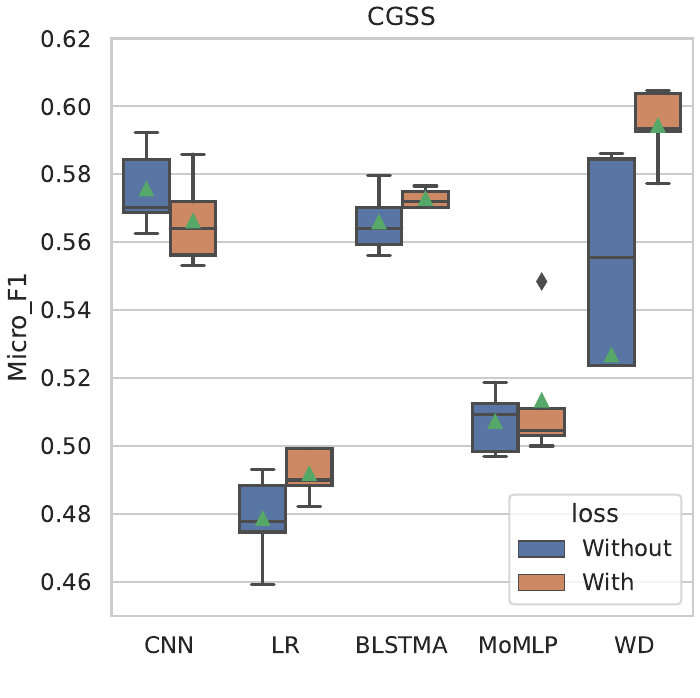} \label{elder}
	}
	\subfloat[\textit{health good on CGSS}]{
		\includegraphics[width=0.24\linewidth]{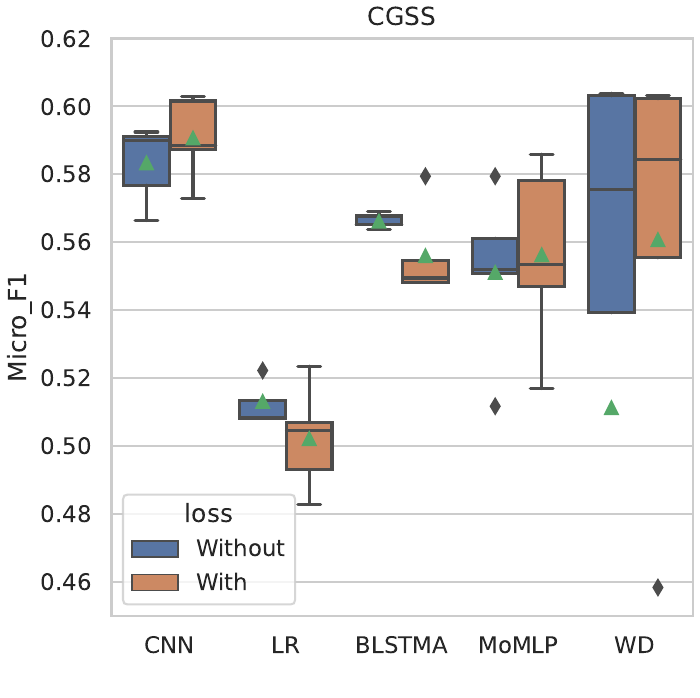} \label{health_good}
	}
	\subfloat[\textit{health bad on CGSS}]{
		\includegraphics[width=0.24\linewidth]{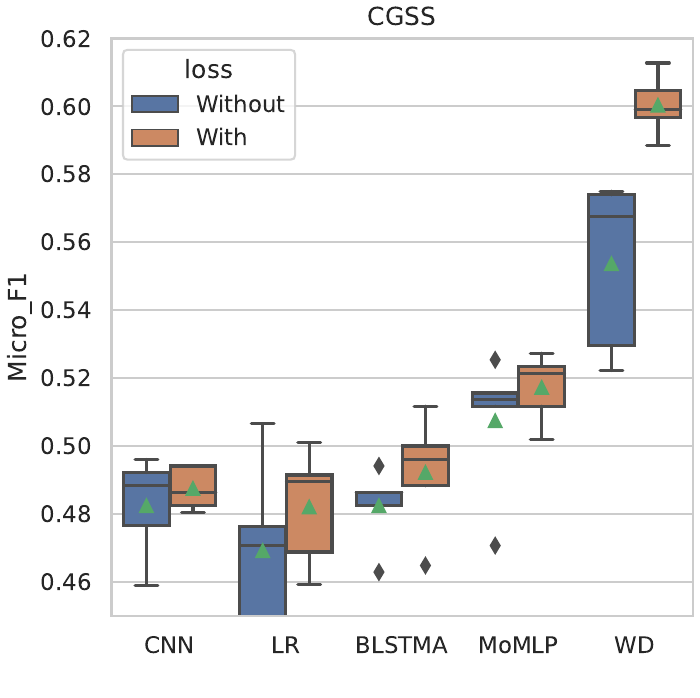} \label{health_bad}
	}
	\caption{The comparison of prediction accuracy stability among multiple models based on four groups of CGSS and ESS datasets. More details of these four groups on ESS can be found in \textit{Appendix \ref{Appendix:accuracy_stability}}.}
	\label{stability_result}
\end{figure*}

\subsection{Experiments}
\label{experiments}

To demonstrate the effectiveness of our solution, we conduct our solution on the previous experimental settings and particularly aim to answer the following two experiment questions (EQ).
\begin{itemize}[leftmargin=*]
	\item \textbf{EQ1:} Can the domain knowledge enhance the happiness computing performance by guiding model training?
	\item \textbf{EQ2:} Can the domain knowledge enhance the explanation consistency of happiness factors?
\end{itemize}

\subsubsection{Domain Knowledge Enhancing Performance (EQ1)}

The accuracy and stability results for each of the models are presented in this section to demonstrate the effectiveness of the domain knowledge.

\paragraph{Domain knowledge improves method accuracies}
To analyze the comprehensive performance of different models with or without domain knowledge, the averages of Macro\_F1 and Micro\_F1 based on $k$-fold cross validation (k=5) are presented in Table~\ref{macro_micro_f1}. This clearly shows that the experiment results guided by domain knowledge can enhance the happiness computing accuracy in 2-fold assessments of Macro\_F1 and Micro\_F1 on most models. More specifically, the results of models with domain knowledge are much better than those of methods without one, with up to 7.33\% (Macro\_F1) and 6.75\% (Micro\_F1) accuracy gains at WideDeep in the \texttt{elder} group of CGSS. Similar conclusions appear in other models regardless of elder, health good, and health bad groups. In a word, the results of the methods with domain knowledge demonstrate more significant improvements than when they are not on two large-scale datasets, which indicates that the domain knowledge could constrain the model training for better accuracy.

\paragraph{Domain knowledge improves accuracy stability} To testify to the performance stability, auxiliary experiments have been implemented herein. As shown in Figure \ref{stability_result}, box plots visually show the distribution of and skewness among our accuracy results by displaying the data quartiles (or percentiles) and averages. Obviously, the methods with domain knowledge are more stable in terms of accuracy (i.e., shorter box length), and there are higher average values of Macro\_F1, as well as Micro\_F1, than without it (green $\triangle$ represents the averages). More specifically, Figure~\ref{young} shows that the results of Macro\_F1 and Micro\_F1 with domain knowledge have more significant stability than those without one on the \texttt{young} group. Similarly, from the results based on other groups (elder, health good, and health bad), we can infer the same conclusions. Therefore, it is easily demonstrated that the method with a domain knowledge constraint could improve the stability of accuracy.

In short, there are significant improvements in accuracy and stability in models with domain knowledge for happiness prediction accuracy, which demonstrates the effectiveness of our method.

\begin{figure}[t]
	\centering
	\includegraphics[width=1.0\linewidth]{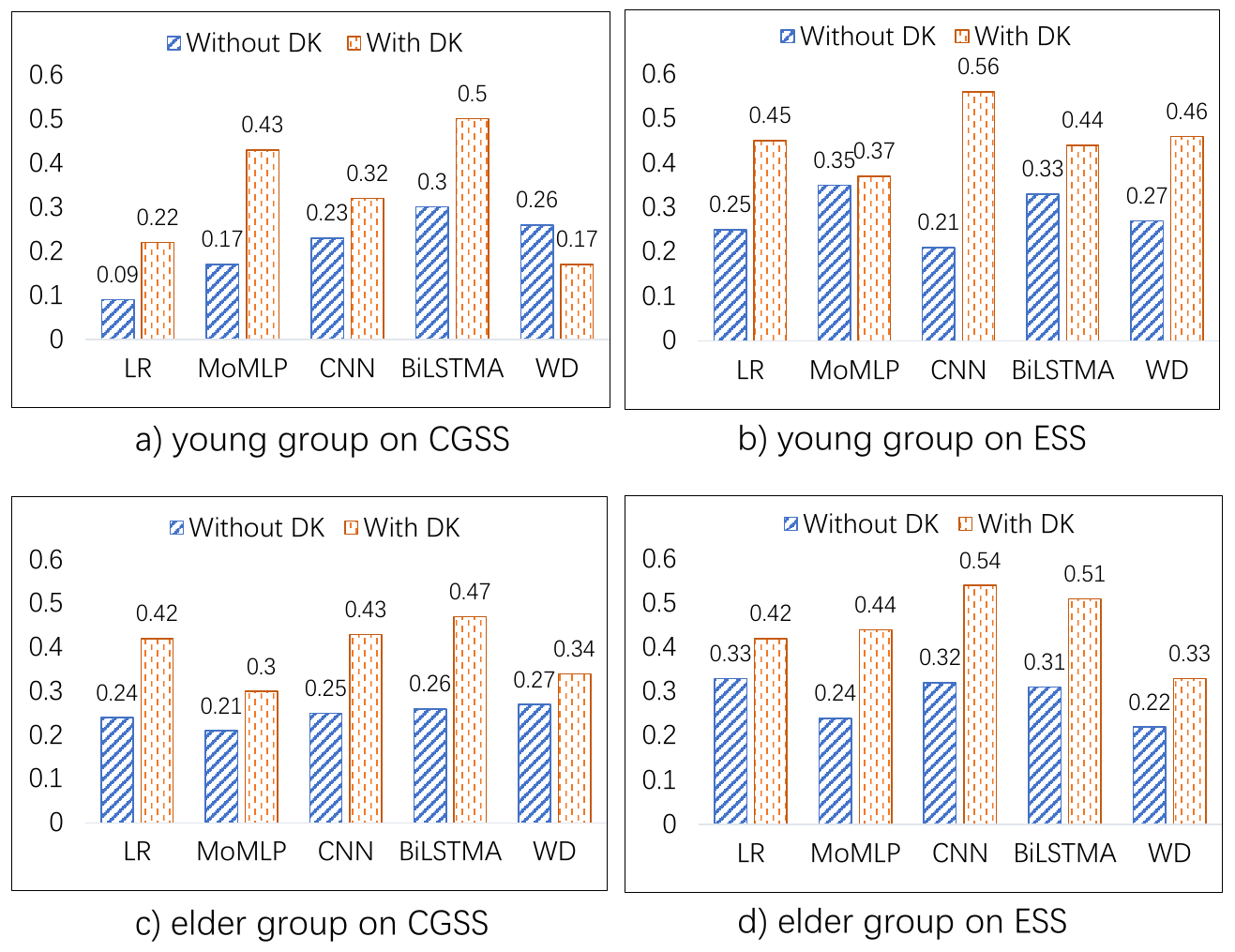}
	\caption{The comparison of Kendall's tau coefficient based on the ranked factor distributions produced by models without or with domain knowledge (DK) on \textit{young} and \textit{elder}  group of CGSS and ESS datasets. More details of other groups can be found in \textit{Appendix \ref{Appendix:kendall_tau}}.}
	\label{kendall_tau_result}
\end{figure}

\subsubsection{Factor Relations Consistency (EQ2)}

Besides the high and stable accuracy, model explanations with factor relations consistency are also crucial to the right prediction for the right reasons. Properties 1 and 2 theoretically infer that there is a relation knowledge of primary and secondary factor relation consistency, which can be combined with importance distribution to constrain the happiness computing model training. In this section, we will show the results constrained by relation knowledge to double demonstrate the consistency of our models with domain knowledge. To evaluate the factor relation consistency, Kendall's tau coefficient is employed in our work for measuring relations between ranked factors.

\begin{table*}[ht]
	\centering
	\tabcolsep=0.9mm
	\caption{The ordinal consistent top-$10$ factors of four groups on CGSS and ESS. Underlines and italics represent the difference between groups built by age (young and elder) and health (health good and health bad), respectively.}
	\label{consistent_factors}
	\begin{tabular}{ccccccccc}
		\toprule
		\multirow{2}{*}{Group} & \multicolumn{4}{c}{CGSS}        & \multicolumn{4}{c}{ESS}     \\
		\cmidrule(r){2-5}         \cmidrule(r){6-9}
		& young  & elder  & health good     & health bad & young & elder  & health good & health bad \\
		\midrule
		1 & equity & status & class & trust & satisfy\_life & satisfy\_life & satisfy\_life & satisfy\_life \\
		2 & family\_status & family\_status & media & income & \underline{social} & \underline{marry} & equity & social \\
		3 & \underline{depression} & \underline{leisure} & public\_service & public\_service & \underline{country} & \underline{work} & \textit{social} & equity \\
		4 & \underline{view} & health & family\_status & media & trust & income & child & marry \\
		5 & trust & house & \textit{work} & family\_status & edu & \underline{health} & politics & edu \\
		6 & \underline{social} & trust & \textit{social} & health & \underline{politics} & trust & \textit{income} & \textit{health} \\
		7 & health & \underline{work} & income & \textit{leisure} & \underline{equity} & country & leisure & \textit{family} \\
		8 & religion & \underline{public\_service} & health & property & income & \underline{leisure} & edu & child \\
		9 & gender & class & depression & \textit{hukou} & age & edu & trust & religion \\
		10 & \underline{income} & \underline{property} & equity & \textit{health\_problem} & \underline{gender} & \underline{family} & health & \textit{public\_service} \\  \bottomrule
	\end{tabular}

\end{table*}

As shown in Figure~\ref{kendall_tau_result}, the averages of Kendall's tau coefficient between the pairwise factor distribution results of models are presented. The results with domain knowledge constraint have significant Kendall's tau coefficient improvements in \texttt{young} and \texttt{elder} groups. On the \texttt{young} group of CGSS dataset, BiLSTMA has the best ordinal consistency improvement up to 0.50, and LR achieves 0.26 improvements gain and up to 0.43. Moreover, for the elder group in the CGSS and ESS datasets, there are 0.35 and 0.22 gains of the CNN with domain knowledge, and up to 0.56 and 0.54, respectively. Similarly, there are significant improvements in ESS dataset, which is present in the deep learning models. The best results are generated by MoMLP, CNN, and BiLSTM, and up to 0.56 in the \texttt{young} group of ESS. In summary, domain knowledge can balance the tension between accuracy and factor explanation, making the right predictions for the right reasons.

\subsection{Implications}

The previous sections discussed the happiness factor list of four specific groups, as shown in Table~\ref{consistent_factors}. Based on these findings, the theoretical and practical implications can improve decision-making.

\subsubsection{Theoretical Implications}
Our findings contribute to the literature on happiness prediction and the quantitative analysis of happiness factors. A deeper examination of the ordinal consistent factors in four groups derived from the datasets deepens our understanding of happiness levels. Specifically, this study contributes to the understanding of ML explainability in a broader sense \cite{Sarthak_2019,Ghanvatkar_2022,Rudin_2023}. Social scientists and policymakers can identify the key factors from 100 wide range of factors for improving human happiness levels. For example, within the \texttt{young} group, \textit{equity, view, trust} and \textit{social} appear in the top-$10$, which indicates that these four have the most significant effects on the happiness of young people. In terms of other factors such as \textit{family\_status}, it influences the \texttt{elder}, \texttt{health good}, and \texttt{health bad} groups, with significance. Similarly, \textit{satisfy\_life} usually reflects the life status of an individual from an indirect perspective and is the most important factor for all groups in the ESS dataset. Therefore, these findings are helpful for research in social science, and our method is also beneficial to transparency improvement in ML research.

\subsubsection{Practical Implications}
There are several vital practical implications. For example, factor relations are references for social decision-making. To verify the rationality of consistent relation factors, we perform a literature review analysis with social science studies to analyze the practical implications.

\begin{itemize}[leftmargin=*]
	\item \textbf{Common factors in groups} For these four groups, the economic factors (e.g., \textit{income, property}), social factors (e.g., \textit{social}), and personal development factors (e.g., \textit{health, health problem, education, class, equity, trust, etc.}) show high importance in happiness level. For the economic factors, numerous works find that family income has a strong impact compared to that of individuals \cite{Oshio_2010,Ambrosio_2020}. Furthermore, concerns of almost all citizens gradually turn toward improving their \textit{health, education}, and \textit{emotion} \cite{Omid_Student_Happiness_2019,Kahriz_2020}. With the continuing development of society, \textit{equity} is increasingly a key factor in view of health equity \cite{Shadmi_2020}, education equity \cite{Ainscow_2020,UNESCO_2017}, etc. These common factors can be classified into two main types including the current emotion and the quality of life, which is consistent with the conclusion by \citet{Kahneman_2006} in the introduction.
	\item \textbf{Difference in groups} However, other factors need to be closely monitored in specific groups. For example, \textit{social, view, depression}, and \textit{education} are strong concerns in the \texttt{young} group, while the \texttt{elder} group focuses more on \textit{health, work, leisure} and \textit{family}. For the \texttt{health good} group, they mainly concentrate on \textit{work, income, social class} and \textit{equity}. In contrast, \textit{health} achieves higher importance in the \texttt{health bad} group \cite{Garaigordobil_2015}.
\end{itemize}

In a word, from this triangulation of evidence, our findings are beneficial to support ML researchers, social scientists, policymakers, and others for making decisions well.
i) For ML researchers, this work provides insights for future research on the explanation of ML; social scientists and policymakers can aim to the key factor set for improving happiness levels. ii) Because ML decisions are opaque, it is not informative enough for policymakers. To this end, this work aims to improve the explanation of ML. By our approach, the happiness computing accuracy and contribution of factors are more reliable to policymakers for decision-making.

\section{Conclusions and Future Work}
It is a challenge that a model not only produces the correct prediction but also arrives at the prediction for the right reasons. To mitigate this, we first conduct an empirical study on various popular models and find the explanation difference even if they are trained on the same datasets and one explanation method. We proved that multiple prediction models have the desirable property of ordinal factor consistency. Then, the factor relations are embedded in importance distribution to guide model training. Our experiments demonstrated that domain knowledge exists and can enhance happiness prediction performance. Moreover, we analyzed the implications of ordinal consistent factors, which verified our findings through double evidence from a literature review using social science evidence. A future direction for our research would be to conduct our method on more online web datasets to further explore the ordinal consistent happiness factors.


\bibliographystyle{ACM-Reference-Format}
\bibliography{www24}


\begin{thebibliography}{56}


\ifx \showCODEN    \undefined \def \showCODEN     #1{\unskip}     \fi
\ifx \showDOI      \undefined \def \showDOI       #1{#1}\fi
\ifx \showISBNx    \undefined \def \showISBNx     #1{\unskip}     \fi
\ifx \showISBNxiii \undefined \def \showISBNxiii  #1{\unskip}     \fi
\ifx \showISSN     \undefined \def \showISSN      #1{\unskip}     \fi
\ifx \showLCCN     \undefined \def \showLCCN      #1{\unskip}     \fi
\ifx \shownote     \undefined \def \shownote      #1{#1}          \fi
\ifx \showarticletitle \undefined \def \showarticletitle #1{#1}   \fi
\ifx \showURL      \undefined \def \showURL       {\relax}        \fi
\providecommand\bibfield[2]{#2}
\providecommand\bibinfo[2]{#2}
\providecommand\natexlab[1]{#1}
\providecommand\showeprint[2][]{arXiv:#2}

\bibitem[Ainscow(2020)]%
        {Ainscow_2020}
\bibfield{author}{\bibinfo{person}{Mel Ainscow}.}
  \bibinfo{year}{2020}\natexlab{}.
\newblock \showarticletitle{Promoting inclusion and equity in education:
  lessons from international experiences}.
\newblock \bibinfo{journal}{\emph{Nordic Journal of Studies in Educational
  Policy}} \bibinfo{volume}{6}, \bibinfo{number}{1} (\bibinfo{year}{2020}),
  \bibinfo{pages}{7--16}.
\newblock


\bibitem[Asai et~al\mbox{.}(2018)]%
        {Akari_2018}
\bibfield{author}{\bibinfo{person}{Akari Asai}, \bibinfo{person}{Sara Evensen},
  \bibinfo{person}{Behzad Golshan}, \bibinfo{person}{Alon Halevy},
  \bibinfo{person}{Vivian Li}, \bibinfo{person}{Andrei Lopatenko},
  \bibinfo{person}{Daniela Stepanov}, \bibinfo{person}{Yoshi Suhara},
  \bibinfo{person}{Wang-chiew Tan}, {and} \bibinfo{person}{Yinzhan Xu}.}
  \bibinfo{year}{2018}\natexlab{}.
\newblock \showarticletitle{HappyDB: A Corpus of 100,000 Crowdsourced Happy
  Moments}.
\newblock  (\bibinfo{year}{2018}).
\newblock


\bibitem[Bolukbasi et~al\mbox{.}(2016)]%
        {Bolukbasi_2016}
\bibfield{author}{\bibinfo{person}{Tolga Bolukbasi}, \bibinfo{person}{Kai{-}Wei
  Chang}, \bibinfo{person}{James~Y. Zou}, \bibinfo{person}{Venkatesh
  Saligrama}, {and} \bibinfo{person}{Adam~Tauman Kalai}.}
  \bibinfo{year}{2016}\natexlab{}.
\newblock \showarticletitle{Man is to Computer Programmer as Woman is to
  Homemaker? Debiasing Word Embeddings}. In \bibinfo{booktitle}{\emph{Advances
  in Neural Information Processing Systems 29: Annual Conference on Neural
  Information Processing Systems 2016, December 5-10, 2016, Barcelona, Spain}}.
  \bibinfo{pages}{4349--4357}.
\newblock


\bibitem[Bublyk et~al\mbox{.}(2022)]%
        {Bublyk2022}
\bibfield{author}{\bibinfo{person}{Myroslava Bublyk}, \bibinfo{person}{Victoria
  Feshchyn}, \bibinfo{person}{Lennara Bekirova}, {and} \bibinfo{person}{Olena
  Khomuliak}.} \bibinfo{year}{2022}\natexlab{}.
\newblock \showarticletitle{Sustainable Development by a Statistical Analysis
  of Country Rankings by the Population Happiness Level}. In
  \bibinfo{booktitle}{\emph{Proceedings of the 6th International Conference on
  Computational Linguistics and Intelligent Systems {(COLINS} 2022). Volume
  {I:} Main Conference, Gliwice, Poland, May 12-13, 2022}}.
  \bibinfo{pages}{817--837}.
\newblock


\bibitem[Chen and Zhang(2013)]%
        {Wanting_2013}
\bibfield{author}{\bibinfo{person}{Wanting Chen} {and} \bibinfo{person}{Xiumei
  Zhang}.} \bibinfo{year}{2013}\natexlab{}.
\newblock \showarticletitle{Analysis of Chinese residents' subjective
  well-being and its influencing factors: based on CGSS2010 data}.
\newblock \bibinfo{journal}{\emph{The world of survey and research}}
  \bibinfo{volume}{10} (\bibinfo{year}{2013}), \bibinfo{pages}{9--15}.
\newblock


\bibitem[Dadvand et~al\mbox{.}(2016)]%
        {Dadvand_2016}
\bibfield{author}{\bibinfo{person}{Payam Dadvand}, \bibinfo{person}{Xavier
  Bartoll}, \bibinfo{person}{Xavier Basaga{\~n}a}, \bibinfo{person}{Albert
  Dalmau-Bueno}, \bibinfo{person}{David Martinez}, \bibinfo{person}{Albert
  Ambros}, \bibinfo{person}{Marta Cirach}, \bibinfo{person}{Margarita
  Triguero-Mas}, \bibinfo{person}{Mireia Gascon}, \bibinfo{person}{Carme
  Borrell}, {and} \bibinfo{person}{Mark~J. Nieuwenhuijsen}.}
  \bibinfo{year}{2016}\natexlab{}.
\newblock \showarticletitle{Green spaces and General Health: Roles of mental
  health status, social support, and physical activity}.
\newblock \bibinfo{journal}{\emph{Environment International}}
  \bibinfo{volume}{91} (\bibinfo{year}{2016}), \bibinfo{pages}{161--167}.
\newblock


\bibitem[D'Ambrosio et~al\mbox{.}(2020)]%
        {Ambrosio_2020}
\bibfield{author}{\bibinfo{person}{Conchita D'Ambrosio},
  \bibinfo{person}{Markus Jäntti}, {and} \bibinfo{person}{Anthony Lepinteur}.}
  \bibinfo{year}{2020}\natexlab{}.
\newblock \showarticletitle{Money and Happiness: Income, Wealth and Subjective
  Well-Being}.
\newblock \bibinfo{journal}{\emph{Social Indicators Research}}
  \bibinfo{volume}{148} (\bibinfo{date}{02} \bibinfo{year}{2020}),
  \bibinfo{pages}{47–66}.
\newblock


\bibitem[Easterlin(2001)]%
        {Richard_2001}
\bibfield{author}{\bibinfo{person}{Richard~A. Easterlin}.}
  \bibinfo{year}{2001}\natexlab{}.
\newblock \showarticletitle{Income and Happiness: Towards a Unified Theory}.
\newblock \bibinfo{journal}{\emph{The Economic Journal}} \bibinfo{volume}{111},
  \bibinfo{number}{473} (\bibinfo{year}{2001}), \bibinfo{pages}{465--484}.
\newblock


\bibitem[Egilmez et~al\mbox{.}(2019)]%
        {Omid_Student_Happiness_2019}
\bibfield{author}{\bibinfo{person}{Gokhan Egilmez},
  \bibinfo{person}{Nadiye~{\"{O}}zlem Erdil}, \bibinfo{person}{Omid~Mohammadi
  Arani}, {and} \bibinfo{person}{Mana Vahid}.} \bibinfo{year}{2019}\natexlab{}.
\newblock \showarticletitle{Application of artificial neural networks to assess
  student happiness}.
\newblock \bibinfo{journal}{\emph{International Journal of Applied Decision
  Sciences}} \bibinfo{volume}{12}, \bibinfo{number}{2} (\bibinfo{year}{2019}),
  \bibinfo{pages}{115--140}.
\newblock


\bibitem[Evensen et~al\mbox{.}(2019)]%
        {Evensen_2019}
\bibfield{author}{\bibinfo{person}{Sara Evensen}, \bibinfo{person}{Yoshihiko
  Suhara}, \bibinfo{person}{Alon~Y. Halevy}, \bibinfo{person}{Vivian Li},
  \bibinfo{person}{Wang{-}Chiew Tan}, {and} \bibinfo{person}{Saran Mumick}.}
  \bibinfo{year}{2019}\natexlab{}.
\newblock \showarticletitle{Happiness Entailment: Automating Suggestions for
  Well-Being}. In \bibinfo{booktitle}{\emph{8th International Conference on
  Affective Computing and Intelligent Interaction, {ACII} 2019, Cambridge,
  United Kingdom, September 3-6, 2019}}. \bibinfo{pages}{62--68}.
\newblock


\bibitem[Garaigordobil(2015)]%
        {Garaigordobil_2015}
\bibfield{author}{\bibinfo{person}{Maite Garaigordobil}.}
  \bibinfo{year}{2015}\natexlab{}.
\newblock \showarticletitle{Predictor variables of happiness and its connection
  with risk and protective factors for health}.
\newblock \bibinfo{journal}{\emph{Frontiers in psychology}}
  \bibinfo{volume}{6} (\bibinfo{date}{08} \bibinfo{year}{2015}),
  \bibinfo{pages}{1176}.
\newblock


\bibitem[Ghanvatkar and Rajan(2022)]%
        {Ghanvatkar_2022}
\bibfield{author}{\bibinfo{person}{Suparna Ghanvatkar} {and}
  \bibinfo{person}{Vaibhav Rajan}.} \bibinfo{year}{2022}\natexlab{}.
\newblock \showarticletitle{Towards a Theory-Based Evaluation of Explainable
  Predictions in Healthcare}. In \bibinfo{booktitle}{\emph{Proceedings of the
  43rd International Conference on Information Systems (ICIS)}}.
\newblock
\urldef\tempurl%
\url{https://aisel.aisnet.org/icis2022/is\_health/is\_health/5}
\showURL{%
\tempurl}


\bibitem[Godbole et~al\mbox{.}(2004)]%
        {Godbole_2004}
\bibfield{author}{\bibinfo{person}{Shantanu Godbole}, \bibinfo{person}{Abhay
  Harpale}, \bibinfo{person}{Sunita Sarawagi}, {and} \bibinfo{person}{Soumen
  Chakrabarti}.} \bibinfo{year}{2004}\natexlab{}.
\newblock \showarticletitle{Document Classification Through Interactive
  Supervision of Document and Term Labels}. In
  \bibinfo{booktitle}{\emph{Knowledge Discovery in Databases: PKDD 2004}},
  Vol.~\bibinfo{volume}{3202}. \bibinfo{pages}{185--196}.
\newblock


\bibitem[Islam and Goldwasser(2020)]%
        {Islam_Goldwasser_Yoga_Happy_2020}
\bibfield{author}{\bibinfo{person}{Tunazzina Islam} {and} \bibinfo{person}{Dan
  Goldwasser}.} \bibinfo{year}{2020}\natexlab{}.
\newblock \showarticletitle{Does Yoga Make You Happy? Analyzing Twitter User
  Happiness using Textual and Temporal Information}. In
  \bibinfo{booktitle}{\emph{2020 {IEEE} International Conference on Big Data
  {(IEEE} BigData 2020), Atlanta, GA, USA, December 10-13, 2020}}.
  \bibinfo{pages}{4241--4249}.
\newblock


\bibitem[Jain and Wallace(2019)]%
        {Sarthak_2019}
\bibfield{author}{\bibinfo{person}{Sarthak Jain} {and}
  \bibinfo{person}{Byron~C. Wallace}.} \bibinfo{year}{2019}\natexlab{}.
\newblock \showarticletitle{Attention is not Explanation}.
\newblock \bibinfo{journal}{\emph{Proceedings of the 2019 Conference on
  Empirical Methods in Natural Language Processing and the 9th International
  Joint Conference on Natural Language Processing, {EMNLP-IJCNLP} 2019, Hong
  Kong, China, November 3-7, 2019}} (\bibinfo{year}{2019}),
  \bibinfo{pages}{3543–3556}.
\newblock


\bibitem[Kahneman and Krueger(2006)]%
        {Kahneman_2006}
\bibfield{author}{\bibinfo{person}{Daniel Kahneman} {and}
  \bibinfo{person}{Alan~B. Krueger}.} \bibinfo{year}{2006}\natexlab{}.
\newblock \showarticletitle{Developments in the Measurement of Subjective
  Well-Being}.
\newblock \bibinfo{journal}{\emph{Journal of Economic Perspectives}}
  \bibinfo{volume}{20}, \bibinfo{number}{1} (\bibinfo{year}{2006}),
  \bibinfo{pages}{3--24}.
\newblock


\bibitem[Kahriz et~al\mbox{.}(2020)]%
        {Kahriz_2020}
\bibfield{author}{\bibinfo{person}{Bahram~Mahmoodi Kahriz},
  \bibinfo{person}{Joanne~L. Bower}, \bibinfo{person}{Francesca M.}, {and}
  \bibinfo{person}{Julia Vogt}.} \bibinfo{year}{2020}\natexlab{}.
\newblock \showarticletitle{Wanting to Be Happy but Not Knowing How: Poor
  Attentional Control and Emotion-Regulation Abilities Mediate the Association
  Between Valuing Happiness and Depression}.
\newblock \bibinfo{journal}{\emph{Journal of Happiness Studies}}
  \bibinfo{volume}{21} (\bibinfo{year}{2020}), \bibinfo{pages}{2583--2601}.
\newblock


\bibitem[Laaksonen(2018)]%
        {Laaksonen_2018}
\bibfield{author}{\bibinfo{person}{Seppo Laaksonen}.}
  \bibinfo{year}{2018}\natexlab{}.
\newblock \showarticletitle{A Research Note: Happiness by Age is More Complex
  than U-Shaped}.
\newblock \bibinfo{journal}{\emph{Journal of Happiness Studies}}
  \bibinfo{volume}{19} (\bibinfo{date}{02} \bibinfo{year}{2018}),
  \bibinfo{pages}{471–482}.
\newblock


\bibitem[Larsen and Eid(2008)]%
        {Randy_2008}
\bibfield{author}{\bibinfo{person}{Randy~J. Larsen} {and}
  \bibinfo{person}{Michael Eid}.} \bibinfo{year}{2008}\natexlab{}.
\newblock \bibinfo{booktitle}{\emph{Ed Diener and the science of subjective
  well-being}}.
\newblock \bibinfo{publisher}{The Guilford Press}, \bibinfo{address}{New York}.
\newblock
\showISBNx{978-1-59385-581-9}


\bibitem[Layard(2005)]%
        {Layard_2005}
\bibfield{author}{\bibinfo{person}{Richard Layard}.}
  \bibinfo{year}{2005}\natexlab{}.
\newblock \showarticletitle{Rethinking Public Economics: The Implications of
  Rivalry and Habit}.
\newblock \bibinfo{journal}{\emph{Economics and Happiness: Framing the
  Analysis}} (\bibinfo{date}{12} \bibinfo{year}{2005}),
  \bibinfo{pages}{147--169}.
\newblock
\showISBNx{9780199286287}
\urldef\tempurl%
\url{https://doi.org/10.1093/0199286280.003.0006}
\showDOI{\tempurl}


\bibitem[Li et~al\mbox{.}(2022)]%
        {ijcai2022_Lilin}
\bibfield{author}{\bibinfo{person}{Lin Li}, \bibinfo{person}{Xiaohua Wu},
  \bibinfo{person}{Miao Kong}, \bibinfo{person}{Dong Zhou}, {and}
  \bibinfo{person}{Xiaohui Tao}.} \bibinfo{year}{2022}\natexlab{}.
\newblock \showarticletitle{Towards the Quantitative Interpretability Analysis
  of Citizens Happiness Prediction}. In \bibinfo{booktitle}{\emph{Proceedings
  of the Thirty-First International Joint Conference on Artificial
  Intelligence, {IJCAI} 2022, Vienna, Austria, 23-29 July 2022}}.
  \bibinfo{pages}{5094--5100}.
\newblock


\bibitem[Lopez-Paz et~al\mbox{.}(2017)]%
        {Lopez_Paz_2017}
\bibfield{author}{\bibinfo{person}{David Lopez-Paz}, \bibinfo{person}{Robert
  Nishihara}, \bibinfo{person}{Soumith Chintala}, \bibinfo{person}{Bernhard
  Schölkopf}, {and} \bibinfo{person}{Léon Bottou}.}
  \bibinfo{year}{2017}\natexlab{}.
\newblock \showarticletitle{Discovering Causal Signals in Images}. In
  \bibinfo{booktitle}{\emph{2017 {IEEE} Conference on Computer Vision and
  Pattern Recognition, {CVPR} 2017, Honolulu, HI, USA, July 21-26, 2017}}.
  \bibinfo{pages}{58--66}.
\newblock
\urldef\tempurl%
\url{https://doi.org/10.1109/CVPR.2017.14}
\showDOI{\tempurl}


\bibitem[Lundberg and Lee(2017)]%
        {Lundberg_2017}
\bibfield{author}{\bibinfo{person}{Scott~M. Lundberg} {and}
  \bibinfo{person}{Su-In Lee}.} \bibinfo{year}{2017}\natexlab{}.
\newblock \showarticletitle{A Unified Approach to Interpreting Model
  Predictions}. In \bibinfo{booktitle}{\emph{Advances in Neural Information
  Processing Systems 30: Annual Conference on Neural Information Processing
  Systems 2017, December 4-9, 2017, Long Beach, CA, {USA}}}.
  \bibinfo{pages}{1--10}.
\newblock


\bibitem[Makridis et~al\mbox{.}(2021)]%
        {Makridis_2021}
\bibfield{author}{\bibinfo{person}{Christos~A. Makridis},
  \bibinfo{person}{David~Y. Zhao}, \bibinfo{person}{Cosmin~A. Bejan}, {and}
  \bibinfo{person}{Gil Alterovitz}.} \bibinfo{year}{2021}\natexlab{}.
\newblock \showarticletitle{Leveraging machine learning to characterize the
  role of socio-economic determinants on physical health and well-being among
  veterans}.
\newblock \bibinfo{journal}{\emph{Computers in Biology and Medicine}}
  \bibinfo{volume}{133} (\bibinfo{year}{2021}), \bibinfo{pages}{104354}.
\newblock
\showISSN{0010-4825}


\bibitem[Musa et~al\mbox{.}(2018)]%
        {Musa_2018}
\bibfield{author}{\bibinfo{person}{Haruna~Danladi Musa},
  \bibinfo{person}{Mohd~Rusli Yacob}, \bibinfo{person}{Ahmad~Makmom Abdullah},
  {and} \bibinfo{person}{Mohd~Yusoff Ishak}.} \bibinfo{year}{2018}\natexlab{}.
\newblock \showarticletitle{Enhancing subjective well-being through strategic
  urban planning: Development and application of community happiness index}.
\newblock \bibinfo{journal}{\emph{Sustainable Cities and Society}}
  \bibinfo{volume}{38} (\bibinfo{year}{2018}), \bibinfo{pages}{184--194}.
\newblock


\bibitem[Oshio et~al\mbox{.}(2011)]%
        {Oshio_2010}
\bibfield{author}{\bibinfo{person}{Takashi Oshio}, \bibinfo{person}{Kayo
  Nozaki}, {and} \bibinfo{person}{Miki Kobayashi}.}
  \bibinfo{year}{2011}\natexlab{}.
\newblock \showarticletitle{Relative Income and Happiness in Asia: Evidence
  from Nationwide Surveys in China, Japan, and Korea}.
\newblock \bibinfo{journal}{\emph{Social Indicators Research}}
  \bibinfo{volume}{104} (\bibinfo{year}{2011}), \bibinfo{pages}{351--367}.
\newblock


\bibitem[Park(2020)]%
        {Park_2020}
\bibfield{author}{\bibinfo{person}{Hyunhee Park}.}
  \bibinfo{year}{2020}\natexlab{}.
\newblock \showarticletitle{{MLP} modeling for search advertising price
  prediction}.
\newblock \bibinfo{journal}{\emph{J. Ambient Intell. Humaniz. Comput.}}
  \bibinfo{volume}{11}, \bibinfo{number}{1} (\bibinfo{year}{2020}),
  \bibinfo{pages}{411--417}.
\newblock


\bibitem[Park et~al\mbox{.}(2022)]%
        {Park_S22}
\bibfield{author}{\bibinfo{person}{Jungin Park}, \bibinfo{person}{Jiyoung Lee},
  \bibinfo{person}{Ig{-}Jae Kim}, {and} \bibinfo{person}{Kwanghoon Sohn}.}
  \bibinfo{year}{2022}\natexlab{}.
\newblock \showarticletitle{Probabilistic Representations for Video Contrastive
  Learning}. In \bibinfo{booktitle}{\emph{2022 IEEE/CVF Conference on Computer
  Vision and Pattern Recognition (CVPR)}}. \bibinfo{publisher}{{IEEE}},
  \bibinfo{pages}{14691--14701}.
\newblock


\bibitem[Peng(2004)]%
        {Peng_2004}
\bibfield{author}{\bibinfo{person}{Danling Peng}.}
  \bibinfo{year}{2004}\natexlab{}.
\newblock \bibinfo{booktitle}{\emph{General Psychology}}.
\newblock \bibinfo{publisher}{Beijing Normal University Press},
  \bibinfo{address}{Beijing, China}.
\newblock


\bibitem[Pérez-Benito et~al\mbox{.}(2019)]%
        {Perez-BenitoVCG19}
\bibfield{author}{\bibinfo{person}{Francisco Pérez-Benito},
  \bibinfo{person}{Patricia Villacampa-Fernández}, \bibinfo{person}{Alberto
  Conejero}, \bibinfo{person}{Juan García-Gómez}, {and}
  \bibinfo{person}{Esperanza Navarro-Pardo}.} \bibinfo{year}{2019}\natexlab{}.
\newblock \showarticletitle{A happiness degree predictor using the conceptual
  data structure for deep learning architectures}.
\newblock \bibinfo{journal}{\emph{Comput. Methods Programs Biomed.}}
  \bibinfo{volume}{168} (\bibinfo{year}{2019}), \bibinfo{pages}{59--68}.
\newblock


\bibitem[Rajendran et~al\mbox{.}(2019)]%
        {Rajendran_2019}
\bibfield{author}{\bibinfo{person}{Arun Rajendran}, \bibinfo{person}{Chiyu
  Zhang}, {and} \bibinfo{person}{Muhammad Abdul-Mageed}.}
  \bibinfo{year}{2019}\natexlab{}.
\newblock \showarticletitle{Happy Together: Learning and Understanding
  Appraisal From Natural Language}. In \bibinfo{booktitle}{\emph{Proceedings of
  the 2nd Workshop on Affective Content Analysis (AffCon 2019) co-located with
  Thirty-Third {AAAI} Conference on Artificial Intelligence {(AAAI} 2019),
  Honolulu, USA, January 27, 2019}}, Vol.~\bibinfo{volume}{2328}.
  \bibinfo{pages}{50--59}.
\newblock


\bibitem[Ribeiro et~al\mbox{.}(2016)]%
        {Ribeiro2016}
\bibfield{author}{\bibinfo{person}{Marco~Tulio Ribeiro},
  \bibinfo{person}{Sameer Singh}, {and} \bibinfo{person}{Carlos Guestrin}.}
  \bibinfo{year}{2016}\natexlab{}.
\newblock \showarticletitle{"Why Should I Trust You?": Explaining the
  Predictions of Any Classifier}. In \bibinfo{booktitle}{\emph{Proceedings of
  the 22nd ACM SIGKDD International Conference on Knowledge Discovery and Data
  Mining}} (San Francisco, California, USA) \emph{(\bibinfo{series}{KDD '16})}.
  \bibinfo{pages}{1135–1144}.
\newblock


\bibitem[Rieger et~al\mbox{.}(2020)]%
        {RiegerSMY2020}
\bibfield{author}{\bibinfo{person}{Laura Rieger}, \bibinfo{person}{Chandan
  Singh}, \bibinfo{person}{W.~James Murdoch}, {and} \bibinfo{person}{Bin Yu}.}
  \bibinfo{year}{2020}\natexlab{}.
\newblock \showarticletitle{Interpretations are Useful: Penalizing Explanations
  to Align Neural Networks with Prior Knowledge}. In
  \bibinfo{booktitle}{\emph{Proceedings of the 37th International Conference on
  Machine Learning, {ICML} 2020, 13-18 July 2020, Virtual Event}}.
  \bibinfo{publisher}{Proceedings of Machine Learning Research},
  \bibinfo{pages}{8116--8126}.
\newblock


\bibitem[Ross et~al\mbox{.}(2017)]%
        {Ross_2017}
\bibfield{author}{\bibinfo{person}{Andrew~Slavin Ross},
  \bibinfo{person}{Michael~C. Hughes}, {and} \bibinfo{person}{Finale
  Doshi-Velez}.} \bibinfo{year}{2017}\natexlab{}.
\newblock \showarticletitle{Right for the Right Reasons: Training
  Differentiable Models by Constraining Their Explanations}. In
  \bibinfo{booktitle}{\emph{Thirty-Fifth {AAAI} Conference on Artificial
  Intelligence, {AAAI} 2021, Thirty-Third Conference on Innovative Applications
  of Artificial Intelligence, {IAAI} 2021, The Eleventh Symposium on
  Educational Advances in Artificial Intelligence, {EAAI} 2021, Virtual Event,
  February 2-9, 2021}} (Melbourne, Australia). \bibinfo{pages}{2662–2670}.
\newblock


\bibitem[Rudin and Shaposhnik(2023)]%
        {Rudin_2023}
\bibfield{author}{\bibinfo{person}{Cynthia Rudin} {and} \bibinfo{person}{Yaron
  Shaposhnik}.} \bibinfo{year}{2023}\natexlab{}.
\newblock \showarticletitle{Globally-Consistent Rule-Based Summary-Explanations
  for Machine Learning Models: Application to Credit-Risk Evaluation}.
\newblock \bibinfo{journal}{\emph{Journal of Machine Learning Research}}
  \bibinfo{volume}{24}, \bibinfo{number}{16} (\bibinfo{year}{2023}),
  \bibinfo{pages}{1--44}.
\newblock


\bibitem[Saputri and Lee(2015)]%
        {Saputri_2015}
\bibfield{author}{\bibinfo{person}{Theresia Saputri} {and}
  \bibinfo{person}{Seok-Won Lee}.} \bibinfo{year}{2015}\natexlab{}.
\newblock \showarticletitle{A Study of Cross-National Differences in Happiness
  Factors Using Machine Learning Approach}.
\newblock \bibinfo{journal}{\emph{International Journal of Software Engineering
  and Knowledge Engineering}}  \bibinfo{volume}{25} (\bibinfo{date}{11}
  \bibinfo{year}{2015}), \bibinfo{pages}{1699--1702}.
\newblock


\bibitem[Schramowski et~al\mbox{.}(2020)]%
        {Schramowski_2020}
\bibfield{author}{\bibinfo{person}{Patrick Schramowski},
  \bibinfo{person}{Wolfgang Stammer}, \bibinfo{person}{Stefano Teso},
  \bibinfo{person}{Anna Brugger}, \bibinfo{person}{Franziska Herbert},
  \bibinfo{person}{Xiaoting Shao}, \bibinfo{person}{Hans{-}Georg Luigs},
  \bibinfo{person}{Anne{-}Katrin Mahlein}, {and} \bibinfo{person}{Kristian
  Kersting}.} \bibinfo{year}{2020}\natexlab{}.
\newblock \showarticletitle{Making deep neural networks right for the right
  scientific reasons by interacting with their explanations}.
\newblock \bibinfo{journal}{\emph{Nat. Mach. Intell.}} \bibinfo{volume}{2},
  \bibinfo{number}{8} (\bibinfo{year}{2020}), \bibinfo{pages}{476--486}.
\newblock


\bibitem[Selvaraju et~al\mbox{.}(2020)]%
        {Selvaraju_2020}
\bibfield{author}{\bibinfo{person}{Ramprasaath~R. Selvaraju},
  \bibinfo{person}{Michael Cogswell}, \bibinfo{person}{Abhishek Das},
  \bibinfo{person}{Ramakrishna Vedantam}, \bibinfo{person}{Devi Parikh}, {and}
  \bibinfo{person}{Dhruv Batra}.} \bibinfo{year}{2020}\natexlab{}.
\newblock \showarticletitle{Grad-CAM: Visual Explanations from Deep Networks
  via Gradient-Based Localization}.
\newblock \bibinfo{journal}{\emph{International Journal of Computer Vision}}
  \bibinfo{volume}{128}, \bibinfo{number}{2} (\bibinfo{year}{2020}),
  \bibinfo{pages}{336–359}.
\newblock
\urldef\tempurl%
\url{https://doi.org/10.1007/s11263-019-01228-7}
\showDOI{\tempurl}


\bibitem[Serrano and Smith(2019)]%
        {Serrano_2019}
\bibfield{author}{\bibinfo{person}{Sofia Serrano} {and}
  \bibinfo{person}{Noah~A. Smith}.} \bibinfo{year}{2019}\natexlab{}.
\newblock \showarticletitle{Is Attention Interpretable?}
\newblock \bibinfo{journal}{\emph{Proceedings of the 57th Conference of the
  Association for Computational Linguistics, {ACL} 2019, Florence, Italy, July
  28- August 2, 2019, Volume 1: Long Papers}} (\bibinfo{year}{2019}),
  \bibinfo{pages}{2931–2951}.
\newblock


\bibitem[Shadmi et~al\mbox{.}(2020)]%
        {Shadmi_2020}
\bibfield{author}{\bibinfo{person}{Efrat Shadmi}, \bibinfo{person}{Yingyao
  Chen}, \bibinfo{person}{In{\^e}s~Costa Dourado}, \bibinfo{person}{Inbal
  Faran-Perach}, \bibinfo{person}{John Furler}, \bibinfo{person}{Peter
  Hangoma}, \bibinfo{person}{Piya Hanvoravongchai}, \bibinfo{person}{Claudia
  Obando}, \bibinfo{person}{Varduhi Petrosyan}, \bibinfo{person}{Krishna~D.
  Rao}, \bibinfo{person}{Ana~Lorena Ruano}, \bibinfo{person}{Leiyu Shi},
  \bibinfo{person}{Luis~Eugenio de Souza}, \bibinfo{person}{Sivan
  Spitzer-Shohat}, \bibinfo{person}{Elizabeth~Ann Sturgiss},
  \bibinfo{person}{Rapeepong Suphanchaimat}, \bibinfo{person}{Manuela~Villar
  Uribe}, {and} \bibinfo{person}{Sara~J. Willems}.}
  \bibinfo{year}{2020}\natexlab{}.
\newblock \showarticletitle{Health equity and COVID-19: global perspectives}.
\newblock \bibinfo{journal}{\emph{International Journal for Equity in Health}}
  \bibinfo{volume}{19}, \bibinfo{number}{1} (\bibinfo{year}{2020}),
  \bibinfo{pages}{104}.
\newblock


\bibitem[Shao et~al\mbox{.}(2021)]%
        {Shao_2021}
\bibfield{author}{\bibinfo{person}{YiJun Shao}, \bibinfo{person}{Ali Ahmed},
  \bibinfo{person}{Angelike~P. Liappis}, \bibinfo{person}{Charles Faselis},
  \bibinfo{person}{Stuart~J. Nelson}, {and} \bibinfo{person}{Qing
  Zeng{-}Treitler}.} \bibinfo{year}{2021}\natexlab{}.
\newblock \showarticletitle{Understanding Demographic Risk Factors for Adverse
  Outcomes in {COVID-19} Patients: Explanation of a Deep Learning Model}.
\newblock \bibinfo{journal}{\emph{Journal of Healthcare Informatics Research}}
  \bibinfo{volume}{5}, \bibinfo{number}{2} (\bibinfo{year}{2021}),
  \bibinfo{pages}{181--200}.
\newblock


\bibitem[Shapley(2016)]%
        {Shapley_2016}
\bibfield{author}{\bibinfo{person}{L.~S. Shapley}.}
  \bibinfo{year}{2016}\natexlab{}.
\newblock \bibinfo{booktitle}{\emph{A Value for n-Person Games}}.
  Vol.~\bibinfo{volume}{2}.
\newblock \bibinfo{publisher}{Princeton University Press},
  \bibinfo{pages}{307--318}.
\newblock


\bibitem[Shrikumar et~al\mbox{.}(2017)]%
        {Shrikumar_2017}
\bibfield{author}{\bibinfo{person}{Avanti Shrikumar}, \bibinfo{person}{Peyton
  Greenside}, {and} \bibinfo{person}{Anshul Kundaje}.}
  \bibinfo{year}{2017}\natexlab{}.
\newblock \showarticletitle{Learning Important Features through Propagating
  Activation Differences}. In \bibinfo{booktitle}{\emph{Proceedings of the 34th
  International Conference on Machine Learning, {ICML} 2017, Sydney, NSW,
  Australia, 6-11 August 2017}} (Sydney, NSW, Australia)
  \emph{(\bibinfo{series}{Proceedings of Machine Learning Research})}.
  \bibinfo{pages}{3145–3153}.
\newblock


\bibitem[Sweeney et~al\mbox{.}(2022)]%
        {Sweeney_2022}
\bibfield{author}{\bibinfo{person}{Colm Sweeney}, \bibinfo{person}{Edel Ennis},
  \bibinfo{person}{Maurice Mulvenna}, \bibinfo{person}{Raymond Bond}, {and}
  \bibinfo{person}{Siobhan O~'~Neill}.} \bibinfo{year}{2022}\natexlab{}.
\newblock \showarticletitle{How Machine Learning Classification Accuracy
  Changes in a Happiness Dataset with Different Demographic Groups}.
\newblock \bibinfo{journal}{\emph{Computers}}  \bibinfo{volume}{11}
  (\bibinfo{date}{05} \bibinfo{year}{2022}), \bibinfo{pages}{83}.
\newblock
\urldef\tempurl%
\url{https://doi.org/10.3390/computers11050083}
\showDOI{\tempurl}


\bibitem[Teso and Kersting(2019)]%
        {Teso_2019}
\bibfield{author}{\bibinfo{person}{Stefano Teso} {and}
  \bibinfo{person}{Kristian Kersting}.} \bibinfo{year}{2019}\natexlab{}.
\newblock \showarticletitle{Explanatory Interactive Machine Learning}. In
  \bibinfo{booktitle}{\emph{Proceedings of the 2019 {AAAI/ACM} Conference on
  AI, Ethics, and Society, {AIES} 2019, Honolulu, HI, USA, January 27-28,
  2019}}. \bibinfo{publisher}{ACM}, \bibinfo{pages}{239--245}.
\newblock


\bibitem[UNESCO(2017)]%
        {UNESCO_2017}
\bibfield{author}{\bibinfo{person}{UNESCO}.} \bibinfo{year}{2017}\natexlab{}.
\newblock \bibinfo{booktitle}{\emph{A Guide for ensuring inclusion and equity
  in education}}.
\newblock \bibinfo{publisher}{Paris:UNESCO}.
\newblock
\showISBNx{978-92-3-100222-9}


\bibitem[Wang et~al\mbox{.}(2019)]%
        {Wang_2019}
\bibfield{author}{\bibinfo{person}{Haohan Wang}, \bibinfo{person}{Zexue He},
  \bibinfo{person}{Zachary~C. Lipton}, {and} \bibinfo{person}{Eric~P. Xing}.}
  \bibinfo{year}{2019}\natexlab{}.
\newblock \showarticletitle{Learning Robust Representations by Projecting
  Superficial Statistics Out}. In \bibinfo{booktitle}{\emph{7th International
  Conference on Learning Representations, {ICLR} 2019, USA, May 6-9, 2019}}.
\newblock


\bibitem[Wang et~al\mbox{.}(2022a)]%
        {cikm_WangLLZ2022}
\bibfield{author}{\bibinfo{person}{Lei Wang}, \bibinfo{person}{Ee{-}Peng Lim},
  \bibinfo{person}{Zhiwei Liu}, {and} \bibinfo{person}{Tianxiang Zhao}.}
  \bibinfo{year}{2022}\natexlab{a}.
\newblock \showarticletitle{Explanation Guided Contrastive Learning for
  Sequential Recommendation}. In \bibinfo{booktitle}{\emph{Proceedings of the
  31st {ACM} International Conference on Information {\&} Knowledge Management,
  Atlanta, GA, USA, October 17-21, 2022}}. \bibinfo{publisher}{{ACM}},
  \bibinfo{pages}{2017--2027}.
\newblock


\bibitem[Wang et~al\mbox{.}(2022b)]%
        {Wang_Weiwei_2022}
\bibfield{author}{\bibinfo{person}{Weiwei Wang}, \bibinfo{person}{Yan Sun},
  \bibinfo{person}{Yong Chen}, \bibinfo{person}{Ya Bu}, {and}
  \bibinfo{person}{Gen Li}.} \bibinfo{year}{2022}\natexlab{b}.
\newblock \showarticletitle{Health Effects of Happiness in China}.
\newblock \bibinfo{journal}{\emph{International journal of environmental
  research and public health}} \bibinfo{volume}{19}, \bibinfo{number}{11}
  (\bibinfo{year}{2022}), \bibinfo{pages}{6686}.
\newblock


\bibitem[Watson et~al\mbox{.}(2023)]%
        {wacv_WatsonHM23}
\bibfield{author}{\bibinfo{person}{Matthew Watson}, \bibinfo{person}{Bashar
  Awwad~Shiekh Hasan}, {and} \bibinfo{person}{Noura~Al Moubayed}.}
  \bibinfo{year}{2023}\natexlab{}.
\newblock \showarticletitle{Learning How to MIMIC: Using Model Explanations to
  Guide Deep Learning Training}. In \bibinfo{booktitle}{\emph{{IEEE/CVF} Winter
  Conference on Applications of Computer Vision, {WACV} 2023, Waikoloa, HI,
  USA, January 2-7, 2023}}. \bibinfo{publisher}{IEEE},
  \bibinfo{pages}{1461--1470}.
\newblock


\bibitem[Xiao et~al\mbox{.}(2021)]%
        {Yuanqing_2021}
\bibfield{author}{\bibinfo{person}{Kai~Yuanqing Xiao}, \bibinfo{person}{Logan
  Engstrom}, \bibinfo{person}{Andrew Ilyas}, {and} \bibinfo{person}{Aleksander
  Madry}.} \bibinfo{year}{2021}\natexlab{}.
\newblock \showarticletitle{Noise or Signal: The Role of Image Backgrounds in
  Object Recognition}. In \bibinfo{booktitle}{\emph{9th International
  Conference on Learning Representations, {ICLR} 2021, Virtual Event, Austria,
  May 3-7, 2021}}.
\newblock


\bibitem[Xin and Inkpen(2019)]%
        {Xin_Inkpen_2019}
\bibfield{author}{\bibinfo{person}{Weizhao Xin} {and} \bibinfo{person}{Diana
  Inkpen}.} \bibinfo{year}{2019}\natexlab{}.
\newblock \showarticletitle{Happiness Ingredients Detection using Multi-Task
  Deep Learning}. In \bibinfo{booktitle}{\emph{Proceedings of the 2nd Workshop
  on Affective Content Analysis (AffCon 2019) co-located with Thirty-Third
  {AAAI} Conference on Artificial Intelligence {(AAAI} 2019), Honolulu, USA,
  January 27, 2019}}, Vol.~\bibinfo{volume}{2328}. \bibinfo{pages}{164--170}.
\newblock


\bibitem[Xue et~al\mbox{.}(2011)]%
        {Xue_2011}
\bibfield{author}{\bibinfo{person}{Yunlian Xue}, \bibinfo{person}{Qirun Niu},
  \bibinfo{person}{Manyun Wang}, \bibinfo{person}{Qiuhua Huang}, {and}
  \bibinfo{person}{Guihao Liu}.} \bibinfo{year}{2011}\natexlab{}.
\newblock \showarticletitle{Analysis of malignant tumors in different age
  groups in our hospital}.
\newblock \bibinfo{journal}{\emph{Chinese Medical Record}}
  \bibinfo{volume}{12}, \bibinfo{number}{2} (\bibinfo{year}{2011}),
  \bibinfo{pages}{49--50}.
\newblock


\bibitem[Young(1985)]%
        {Young_1985}
\bibfield{author}{\bibinfo{person}{H.~P. Young}.}
  \bibinfo{year}{1985}\natexlab{}.
\newblock \showarticletitle{Monotonic solutions of cooperative games}.
\newblock \bibinfo{journal}{\emph{International Journal of Game Theory}}
  \bibinfo{volume}{14} (\bibinfo{year}{1985}), \bibinfo{pages}{65--72}.
\newblock
\urldef\tempurl%
\url{https://doi.org/10.1007/BF01769885}
\showDOI{\tempurl}


\bibitem[Yu and Wang(2017)]%
        {Yu_2017}
\bibfield{author}{\bibinfo{person}{Zonghuo Yu} {and} \bibinfo{person}{Fei
  Wang}.} \bibinfo{year}{2017}\natexlab{}.
\newblock \showarticletitle{Income Inequality and Happiness: An Inverted
  U-Shaped Curve}.
\newblock \bibinfo{journal}{\emph{Frontiers in Psychology}}
  \bibinfo{volume}{8} (\bibinfo{date}{11} \bibinfo{year}{2017}),
  \bibinfo{pages}{2052}.
\newblock


\bibitem[Zhang et~al\mbox{.}(2021)]%
        {Cui_Peng_2021}
\bibfield{author}{\bibinfo{person}{Xingxuan Zhang}, \bibinfo{person}{Peng Cui},
  \bibinfo{person}{Renzhe Xu}, \bibinfo{person}{Linjun Zhou},
  \bibinfo{person}{Yue He}, {and} \bibinfo{person}{Zheyan Shen}.}
  \bibinfo{year}{2021}\natexlab{}.
\newblock \showarticletitle{Deep Stable Learning for Out-Of-Distribution
  Generalization}. In \bibinfo{booktitle}{\emph{{IEEE} Conference on Computer
  Vision and Pattern Recognition, {CVPR} 2021, virtual, June 19-25, 2021}}.
  \bibinfo{pages}{5368--5378}.
\newblock


\end{thebibliography}

\newpage

\setcounter{table}{0}
\setcounter{figure}{0}

\renewcommand{\thetable}{A\arabic{table}}
\renewcommand{\thefigure}{A\arabic{figure}}

\begin{appendices}

	\section{Factor Introduction}
	\label{Appendix:Factors_Introduction}

	Our experiments are based on the public Chinese General Social Survey (CGSS) \footnote{http://cgss.ruc.edu.cn/} 2015 (full edition) and European Social Survey (ESS) \footnote{https://ess-search.nsd.no/} 2018 datasets. The descriptions of key factors are introduced as follows, and the entire factor details can be found on the corresponding web page.

	\paragraph{CGSS}
	Chinese General Social Survey (CGSS), started in 2003, is the earliest nationwide, comprehensive, and continuous academic survey project in China. CGSS systematically and comprehensively collects data at multiple levels of society, community, family, and individual, summarizes the trend of social change, discusses issues of great scientific and practical significance, promotes the openness and sharing of domestic scientific research, provides data for international comparative research, and acts as a multidisciplinary economic and social data collection platform. At present, CGSS data has become the most important data source for the study of Chinese society and is widely used in scientific research, teaching, and government decision-making.

	In this section, the factors that appear in this paper are introduced below, more details can be found on the website.

	\paragraph{ESS}
	The European Social Survey (ESS) is an academically-driven multi-country survey, which has been administered in over 30 countries to date. Its three aims are, firstly, to monitor and interpret changing public attitudes and values within Europe and to investigate how they interact with Europe's changing institutions, secondly, to advance and consolidate improved methods of cross-national survey measurement in Europe and beyond, and thirdly, to develop a series of European social indicators, including attitudinal indicators.

	The survey involves strict random probability sampling, a minimum target response rate of 70\%, and rigorous translation protocols. The hour-long face-to-face interview includes questions on a variety of core topics repeated from previous rounds of the survey and also two modules developed for Round 8 covering Public Attitudes to Climate Change, Energy Security, and Energy Preferences and Welfare Attitudes in a Changing Europe (the latter is a partial repeat of a module from Round 4).

	\section{Properties of Shapley Value}
	\label{Appendix:Properties_of_Shapley}

	\begin{table*} \small
		\centering
		\tabcolsep=0.7mm
		\renewcommand\arraystretch{1.0}
		\caption{The results of all methods with or without happiness domain knowledge of health good and health bad groups on CGSS and ESS datasets.}
		\label{Appendix:macro_micro_f1}
		\begin{tabular}{ccccccccccccc}
			\toprule
			\multirow{2}{*}{Group}       & \multirow{2}{*}{Dataset} & \multicolumn{1}{c}{\multirow{2}{*}{\makecell{Domain \\ Knowledge}}} & \multicolumn{2}{c}{CNN[2019]}       & \multicolumn{2}{c}{LR[2015]}      & \multicolumn{2}{c}{BiLSTMA[2020]}                                      & \multicolumn{2}{c}{MoMLP[2020]}             & \multicolumn{2}{c}{WideDeep [2022]}  \\
			\cmidrule(r){4-5}		 \cmidrule(r){6-7}		 \cmidrule(r){8-9}	 \cmidrule(r){10-11}  \cmidrule(r){12-13}
			&                          & \multicolumn{1}{c}{}                                                  & \multicolumn{1}{c}{Macro\_F1} & \multicolumn{1}{c}{Micro\_F1} & \multicolumn{1}{c}{Macro\_F1} & \multicolumn{1}{c}{Micro\_F1} & \multicolumn{1}{c}{Macro\_F1} & \multicolumn{1}{c}{Micro\_F1} & \multicolumn{1}{c}{Macro\_F1} & \multicolumn{1}{c}{Micro\_F1} & \multicolumn{1}{c}{Macro\_F1} & \multicolumn{1}{c}{Micro\_F1}  \\
			\midrule
			\multirow{4}{*}{\makecell{Health\\ Good}}       & \multirow{2}{*}{CGSS}    & \XSolidBrush
			& 63.11 & 58.33 & \textbf{47.50} & \textbf{51.31} & \textbf{61.02} & \textbf{56.61} & 52.97 & 55.10 & 51.24 & 51.12 \\
			&   	& \checkmark
			& \textbf{63.41} & \textbf{58.70} & 46.54 & 50.22 & 60.24 & 55.60 & \textbf{55.02} & \textbf{55.63} & \textbf{56.09} & \textbf{56.07} \\
			& \multirow{2}{*}{ESS}     & \XSolidBrush
			& 63.50 & 59.06 & 30.09 & 40.86 & 67.95 & 65.15 & 66.04 & 66.27 & 65.68 & 65.61 \\
			&   	& \checkmark
			& \textbf{65.81} & \textbf{64.93} & \textbf{30.95} & \textbf{41.75} & \textbf{69.98} & \textbf{66.76} & \textbf{69.11} & \textbf{68.97} & \textbf{66.65} & \textbf{66.67} \\
			\multirow{4}{*}{\makecell{Health \\ Bad}}       & \multirow{2}{*}{CGSS}    & \XSolidBrush
			& 49.11 & 48.24 & 40.25 & 46.92 & 47.85 & 48.24 & 41.78 & 50.74 & 54.51 & 55.37 \\
			&   	& \checkmark
			& \textbf{49.68} & \textbf{48.75} & \textbf{41.46} & \textbf{48.20} & \textbf{48.39} & \textbf{49.22} & \textbf{43.79} & \textbf{51.72} & \textbf{60.10} & \textbf{60.03} \\
			& \multirow{2}{*}{ESS}     & \XSolidBrush
			& 49.66 & 51.53 & 29.42 & 36.07 & \textbf{52.07} & \textbf{51.69} & 37.37 & 49.54 & 65.79 & 65.70 \\
			&   	& \checkmark
			& \textbf{50.12} & \textbf{51.81} & \textbf{30.22} & \textbf{36.91} & 51.72 & 51.24 & \textbf{38.66}	& \textbf{49.71}  & \textbf{66.90} & \textbf{66.93} \\
			\bottomrule
		\end{tabular}
	\end{table*}

	\begin{figure*}[ht]
		\centering
		\subfloat[\textit{young}]{
			\includegraphics[width=0.22\linewidth]{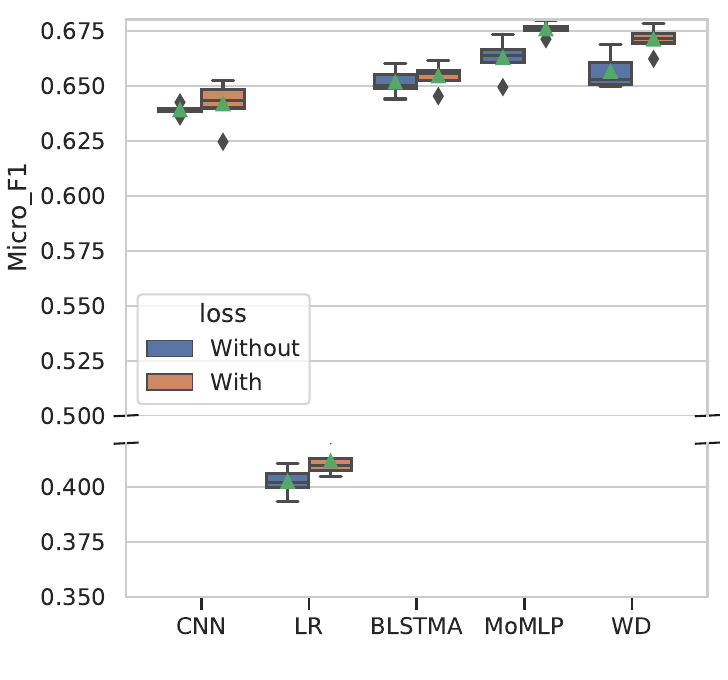}
		}
		\subfloat[\textit{elder}]{
			\includegraphics[width=0.22\linewidth]{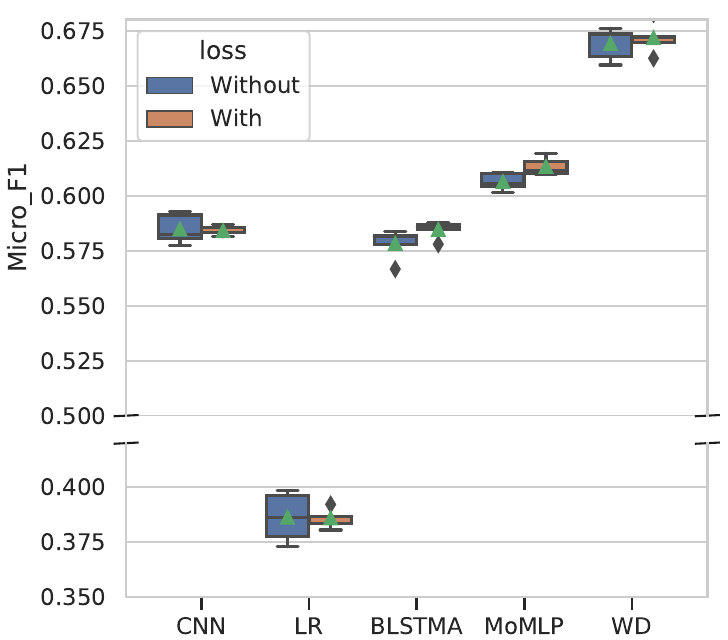}
		}
		\subfloat[\textit{health good}]{
			\includegraphics[width=0.22\linewidth]{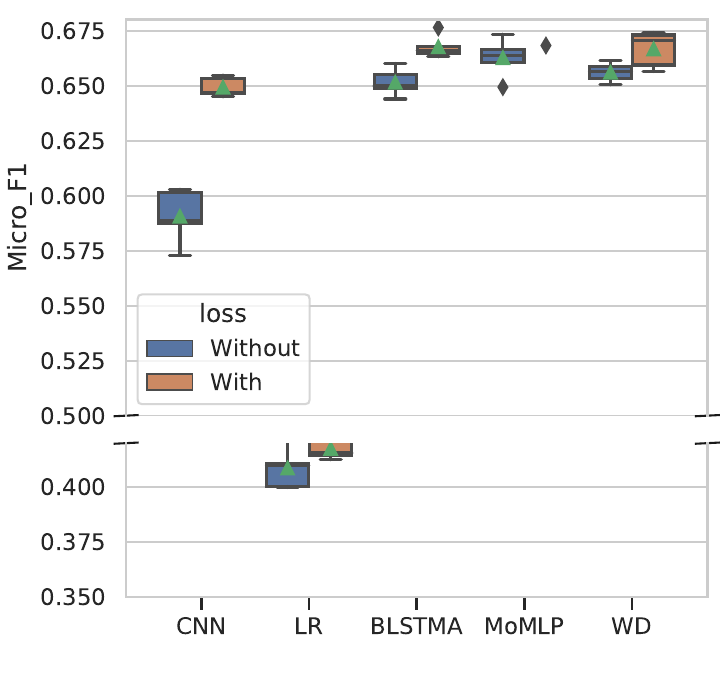}
		}
		\subfloat[\textit{health bad}]{
			\includegraphics[width=0.22\linewidth]{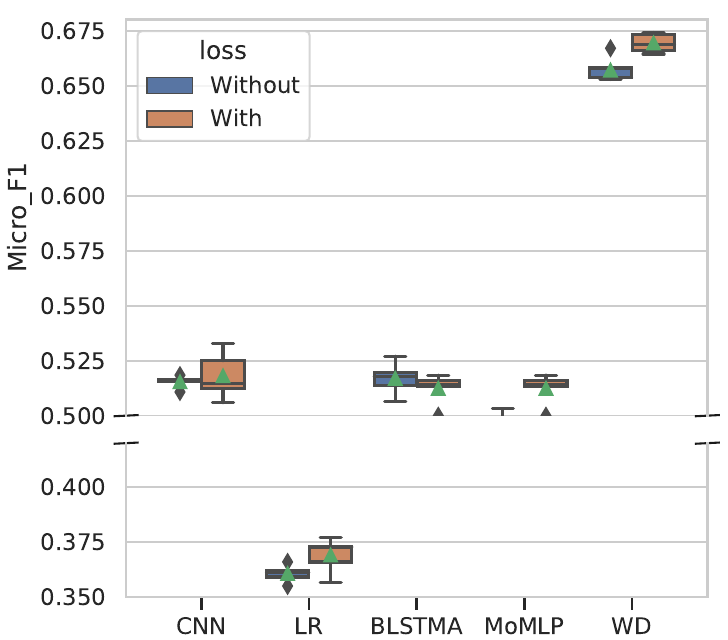}
		}
		\caption{The comparison of prediction accuracy stability among multiple models based on four groups of ESS datasets.}
		\label{Appendix:stability}
	\end{figure*}

	Moreover, the Shapley value has the following desirable properties:

	\begin{enumerate}[start=1, leftmargin=*]
		\item \textit{Efficiency: The total gain is distributed:
			\begin{equation}
				v(\mathcal{S}) = \phi_{0} + \sum\nolimits_{j=1}^{S} \phi_{j},
			\end{equation}
			where $\mathcal{S} \subseteq \mathcal{M} \setminus \{j\}$ is a subset consisting of $|\mathcal{S}|$ factors, and $\phi_0$ denotes the base value.}
		\item \textit{Symmetry: If $i$ and $j$ are two players who contribute equally to all possible coalitions, i.e.
			\begin{equation}
				v(\mathcal{S} \cup \{i\}) = v(\mathcal{S} \cup \{j\})
			\end{equation}
			for every subset $S$ which contains neither $i$ nor $j$, then their Shapley values are identical: $\phi_{i} = \phi_{j}$.}

		\item \textit{Dummy: If $v(\mathcal{S} \cup \{j\}) = v(\mathcal{S})$ for a player $j$ and all coalitions $\mathcal{S}$, then $\phi_{j}=0$.}

		\item \textit{Linearity: If two coalition games described by gain functions $v$ and $w$ are combined, then the distributed gains correspond to the gains derived from $v$ and the gains derived from $w$:
			\begin{equation}
				\phi_{i}(v + w) = \phi_{i}(v) + \phi_{i}(w)
			\end{equation}
			for every $i$. Also, for any real number, we have that
			\begin{equation}
				\phi_{i}(av) = a\phi_{i}(v)
			\end{equation}
		}
	\end{enumerate}

	\section{Prediction Accuracy on Health Good and Health Bad Groups}
	\label{Appendix:F1_Results_health}

	The Macro\_F1 and Micro\_F1 of our models without and with domain knowledge on health good and health bad groups are shown in Table~\ref{Appendix:macro_micro_f1}. The WideDeep achieves the best improvements upon to 5.59\% (health good group on CGSS) and 4.95\% (health bad group on CGSS) on Macro\_F1 and Micro\_F1 respectively. As the young and elder groups, it demonstrated that there are significant improvements in the models with domain knowledge.

\begin{figure*}[ht]
	\centering
	\subfloat[without DK on health good group \\ of CGSS]{ 		\includegraphics[width=0.22\linewidth]{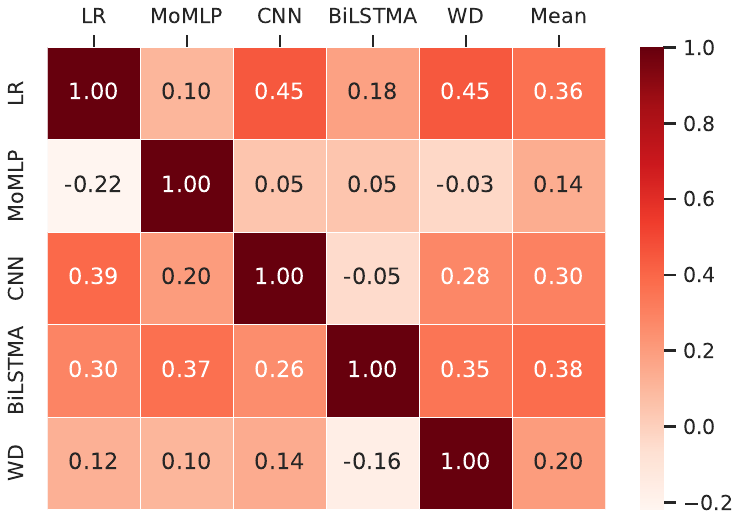}
			\label{CGSS_healthGood_kendall_tau}
		}
	\subfloat[with DK on health good group \\ of CGSS]{
			\includegraphics[width=0.22\linewidth]{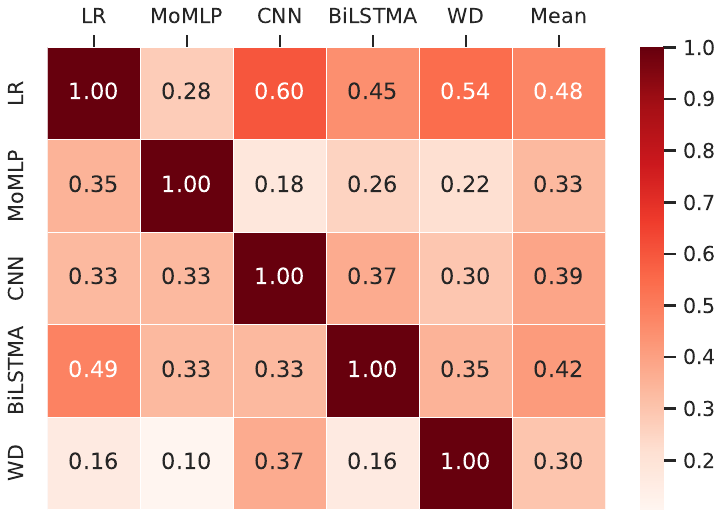}
			\label{CGSS_healthGood_KL_kendall_tau}
		}
	\subfloat[without DK on health bad group \\ of CGSS]{ 		\includegraphics[width=0.22\linewidth]{result_figure/CGSS_kendall_tau/CGSS_healthBad_kendall_tau.pdf}
			\label{CGSS_healthBad_kendall_tau}
		}
	\subfloat[with DK on health bad group \\ of  CGSS]{
			\includegraphics[width=0.22\linewidth]{result_figure/CGSS_kendall_tau/CGSS_healthBad_KL_kendall_tau.pdf}
			\label{CGSS_healthBad_KL_kendall_tau}
		} \\
	\subfloat[without DK on health good group \\ of  ESS]{
			\includegraphics[width=0.22\linewidth]{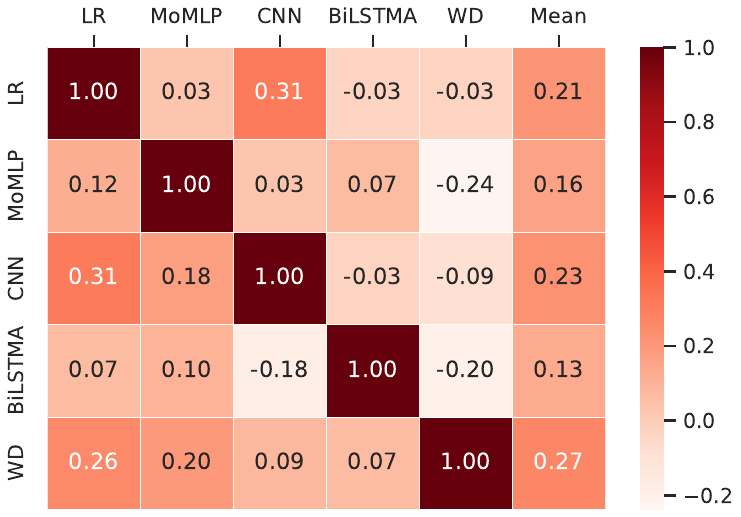}
			\label{ESS_healthGood_kendall_tau}
		}
	\subfloat[with DK on health good group \\ of ESS]{
			\includegraphics[width=0.22\linewidth]{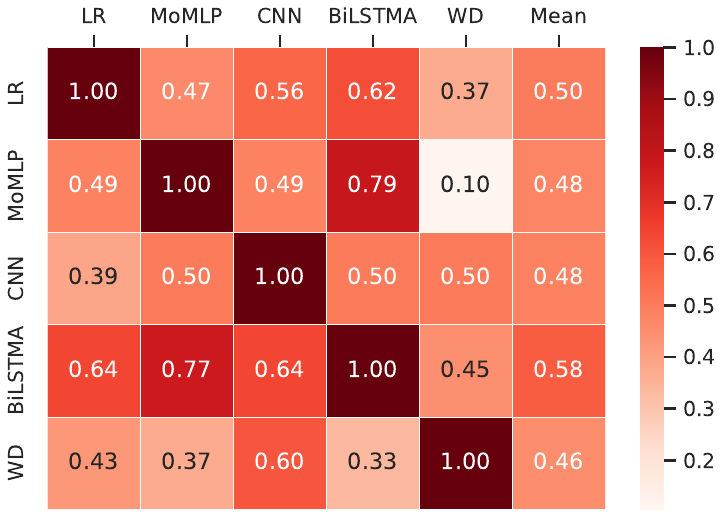}
			\label{ESS_healthGood_KL_kendall_tau}
		}
	\subfloat[without DK on health bad group \\ of  ESS]{
			\includegraphics[width=0.22\linewidth]{result_figure/ESS_kendall_tau/ESS_healthBad_kendall_tau.pdf}
			\label{ESS_healthBad_kendall_tau}
		}
	\subfloat[with DK on health bad group \\ of  ESS]{
			\includegraphics[width=0.22\linewidth]{result_figure/ESS_kendall_tau/ESS_healthBad_KL_kendall_tau.pdf}
			\label{ESS_healthBad_KL_kendall_tau}
		}
	\caption{The comparison of Kendall's tau coefficient based on the ranked factor distributions produced by models without or with domain knowledge (DK) on \textit{health good} and \textit{health bad}  group of CGSS and ESS datasets. }
	\label{Appendix:kendall_tau_result_health}
\end{figure*}

	As shown in Figure~\ref{Appendix:stability}, the prediction stability of our models on four groups of ESS dataset is represented by the file box. Obviously, the prediction of models with domain knowledge gains better stability than that without domain knowledge. It can also indicate the effectiveness of the domain knowledge constraint. For overall accuracy performance, the results with domain knowledge are also better than others based on the averages of Micro\_F1 (represented by green $\triangle$). In addition, we can find that the results of ESS are usually better than those of CGSS (shown in section \ref{experiments}), which is maybe due to the larger amount of training samples.

	\section{Ordinal Factor Consistency in Other Groups}
	\label{Appendix:kendall_tau}

	We can draw the similar conclusion in \texttt{health good} and \texttt{health bad} group. Especially, on the \texttt{health bad} group of CGSS dataset, LR, CNN, and BiLSTMA have a competitive and highly ordinal consistency improvement. Specifically, there are 0.12 (LR) and 0.19 (MoMLP) gains in the models with domain knowledge on the young group of CGSS. For the \texttt{health bad} group of ESS, BiLSTMA has the best performance over others based on the average, and the consistency imprisonments of factor relation up to 0.45. Moreover, LR, CNN, BiLSTMA, and WideDeep show concurrently superior performance on these two datasets. This indicates that the domain knowledge constraint could benefit the model training for learning more consistent factor relations, which is good for right prediction for right reasons.

	In summary, these results indicate that the domain knowledge constraint could benefit the modeling training for learning more consistent factor relations, which is significant for prediction consistency.

\end{appendices}

\end{document}